\documentclass{article}

\usepackage{arxiv}
\usepackage[utf8]{inputenc} 
\usepackage[T1]{fontenc}    
\usepackage{hyperref}       
\usepackage{url}            
\usepackage{booktabs}       
\usepackage{amsfonts}       
\usepackage{nicefrac}       
\usepackage{microtype}  
\usepackage{graphicx}
\usepackage[numbers]{natbib}
\usepackage{doi}
\usepackage{subcaption}
\usepackage{amsmath} 
\usepackage{amsthm}
\usepackage{multirow}
\usepackage{algorithmic} 
\usepackage{algorithm} 
\newtheorem{theorem}{Theorem} 
\newtheorem{assumption}{Assumption}
\usepackage{enumitem}

\title{Gaussian Process Regression with Soft Inequality and  Monotonicity Constraints}
	
\author{ Didem Kochan \\
	Department of Industrial and Systems Engineering\\
	Lehigh University\\
	Bethlehem, PA 18015 \\
	\texttt{dik318@lehigh.edu} \\
	\And
	Xiu Yang \\
	Department of Industrial and Systems Engineering\\
	Lehigh University\\
	Bethlehem, PA 18015 \\
	\texttt{xiy518@lehigh.edu} \\
}

\date{}

\renewcommand{\shorttitle}{GP with Soft Constraints}
\hypersetup{
pdftitle={Gaussian Process Regression with Soft Inequality and Monotonicity Constraints},
pdfsubject={q-bio.NC, q-bio.QM},
pdfauthor={Didem Kochan, Xiu Yang},
pdfkeywords={First keyword, Second keyword, More},
}

\begin{document}
\maketitle
\begin{abstract}
Gaussian process (GP) regression is a non-parametric, Bayesian framework to approximate complex models. Standard GP regression can lead to an unbounded model in which some points can take infeasible values. We introduce a new GP method that enforces the physical constraints in a probabilistic manner. This GP model is trained by the quantum-inspired Hamiltonian Monte Carlo (QHMC). QHMC is an efficient way to sample from a broad class of distributions. Unlike the standard Hamiltonian Monte Carlo algorithm in which a particle has a fixed mass, QHMC allows a particle to have a random mass matrix with a probability distribution. Introducing the QHMC method to the inequality and monotonicity constrained GP regression in the probabilistic sense, our approach improves the accuracy and reduces the variance in the resulting GP model. According to our experiments on several datasets, the proposed approach serves as an efficient method as it accelerates the sampling process while maintaining the accuracy, and it is applicable to high dimensional problems.
\end{abstract}


\section{Introduction}
In many real-world applications, measuring complex systems or evaluating computational models can be time-consuming, costly or computationally intensive. Gaussian process regression~(GPR) is one of the Bayesian techniques that addresses this problem by building a surrogate model. It is a supervised machine learning framework that has been widely used in regression and classification tasks. A GP can be interpreted as a suitable probability distribution on a set of functions, which can be conditioned on observations using Bayes’ rule~\citep{lange2021linearly}. A GP regression model can make predictions incorporating prior information~(kernels) and generate uncertainty measures over predictions~\citep{rasmussen2006gaussian}. However, prior knowledge often includes physical laws, and using the standard GP regression framework may lead to an unbounded model in which some points can take infeasible values that violate physical laws~\citep{lange2021linearly}. For example, non-negativity is a requirement for various physical properties such as temperature, density and viscosity~\citep{nonnegativityXiu}. Therefore, incorporating physical information in GP framework can regularize the behaviour of the model and provide more realistic uncertainties~\citep{swiler2020survey}. 

A significant amount of research has been conducted to incorporate physical information in GP framework, resulting in various techniques and methodologies~\citep{swiler2020survey}. For example, a probit model for the likelihood of derivative information can be employed to enforce monotonicity constraints~\citep{riihimaki2010derivative}. Although this approach can also be used to enforce convexity in one dimension, an additional requirement on Hessian is incorporated for higher dimensions~\citep{convexity2012gaussian}. In~\citep{lopezmonotonicity} an additive GP approach is introduced to account for monotonicity constraints. Although posterior sampling step can be challenging, the additive GP framework enables to satisfy the constraints everywhere in the input space, and it is scalable to higher dimensions.

Enforcing inequality constraints into a GP, on the other hand, is more challenging, as the conditional process, subject to these constraints, does not retain the properties of a GP~\citep{maatouk2017gaussian}. One of the approaches to handle this problem is a data augmentation approach in which the inequality constraints are enforced at various locations and approximate samples are drawn from the predictive distribution~\citep{abrahamsen2001kriging}, or using a block covariance kernel~\citep{raissi2017covkernel}. Implicitly constrained GP regression method proposed in~\citep{salzmann2010implicitly} shows
that the mean prediction of a GP implicitly satisfies linear constraints, if the constraints are satisfied
by the training data. A similar approach shows that when we impose linear inequality constraints on a finite set of points in the domain, the resulting process is a compound Gaussian Process with a truncated Gaussian mean~\citep{agrell2019gaussian}.  

Most of the approaches assume that the inequalities are satisfied on a finite set of input locations. Based on that assumption, the methods approximate the posterior distribution given those constraint input points. The approach introduced in~\citep{convexity2012gaussian} is an example of these methods, where maximum likelihood estimation of GP hyperparameters are investigated  under the constraint assumptions. In practice, this should also limit the number of constraint points needed for an effective discrete-location approximation. In addition, the method is not efficient on high-dimensional datasets as it takes a large amount of time to train the GP model. 

The first Gaussian method that satisfies certain inequalities at all the input space is proposed by Maatouk and Bay~\citep{maatouk2017gaussian}. The GP approximation of the samples are performed in the finite-dimensional space functions, and a rejection sampling method is used for approximating the posterior. The convergence properties of the method is investigated in~\citep{maatouk2015cross}. Although using the rejection sampling to obtain posterior helps convergence, it might be computationally expensive. Similar to the previous approaches in which a set of inputs satisfy the constraints, this method also suffers from the curse of dimensionality. 
Later, the truncated Gaussian approach~\citep{lopez2018finite} extends the framework in~\citep{maatouk2017gaussian} to general sets of linear inequalities. Building upon the approaches in~\citep{maatouk2017gaussian} and~\citep{maatouk2015cross}, the work presented in~\citep{lopez2018finite} introduces a finite-dimensional approach that incorporates inequalities for both data interpolation and covariance parameter estimation. In this work, the posterior distribution is expressed as a truncated multinormal distribution. The method uses different Markov Chain Monte Carlo~(MCMC) methods and exact sampling methods to obtain the posterior distribution. Among the various MCMC sampling techniques including Gibbs, Metropolis-Hastings~(MH) and Hamiltonian Monte Carlo~(HMC), the results indicate that HMC sampling is the most efficient one. The truncated Gaussian approaches offer several advantages, including the ability to achieve high accuracy and the flexibility in satisfying multiple inequality conditions. However, although those types of methods address the limitations in~\citep{maatouk2017gaussian}, they might be time consuming particularly in applications with large datasets or high-dimensional spaces. 

In this work, we use QHMC algorithm to train the GP model, and enforce the inequality and monotonicity constraints in a probabilistic manner. Our work addresses the computational limitations caused by high dimensions or large datasets. Unlike truncated Gaussian methods in~\citep{lopez2018finite} for inequality constraints, or additive GP~\citep{lopezmonotonicity} with monotonicity constraints, the proposed method can maintain its efficiency on higher dimensions. Further, we adopt an adaptive learning algorithm that selects the constraint locations. The efficiency and accuracy of the QHMC algorithms are demonstrated on inequality and monotonicity constrained problems. Inequality constrained examples include lower and higher dimensional synthetic problems, a conservative tracer distribution from sparse tracer concentration measurements and  a three-dimensional heat transfer problem, while monotonicity constrained examples provide lower and higher dimensional synthetic problems. Our contributions can be summarized in three key points: (i) QHMC reduces difference between posterior mean and the ground truth, (ii) utilizing QHMC in a probabilistic sense decreases variance and uncertainty, and (iii) the proposed algorithm is a robust, efficient and flexible method applicable to a wide range of problems. We implemented QHMC sampling in the truncated Gaussian approach to enhance accuracy and efficiency while working with the QHMC algorithm.
\section{Gaussian Process under inequality constraints}
\subsection{Standard GP regression framework}
Suppose we have a target function represented by values $\mathbf{y} = (y^{(1)}, y^{(2)}, ..., y^{(T)})^{N}$, where $y^{(i)} \in \mathbb{R}$ are observations at locations $\mathbf{X}= \{x^{(i)}\}_{i=1}^{N}$. Here, $x^{(i)}$ represents $d$-dimensional vectors in the domain $\mathcal{D} \in \mathbb{R}^{d}$. Using the framework provided in~\citep{GPframework}, we approximate the target function by a GP, denoted as $Y(.,.): D \times \Omega \rightarrow \mathrm{R}$. We can express $Y$ as
\begin{align*}
    Y(x) := GP[\mu(x), K(x,x')],
\end{align*}
where $\mu(.)$ is the mean function and $K(x,x')$ is the covariance function defined as
\begin{equation*}
    \mu(x) = \mathbb{E}[Y(x)], \quad \text{and} \quad K(x,x') = \mathbb{E}[Y(x)-\mu(x)][Y(x')-\mu(x')]
\end{equation*}

Typically, the standard squared exponential
covariance kernel can be used as a kernel function:
\begin{equation*}
    K(x,x') = \sigma^{2}\exp\left(-\frac{||x- x'||_{2}^{2}}{2l^{2}} \right) +\sigma^{2}_{n}\delta_{x,x'},
\end{equation*}
where $\sigma^{2}$ is the signal variance, $\delta_{x,x'}$ is the Kronecker delta function and $l$ is the length-scale. We then assume that the observation includes an additive independent identically distributed (i.i.d.) Gaussian noise term $\epsilon$ and having zero mean and variance $\sigma^{2}_{n}$. We denote the hyperparameters by $\mathbf{\theta} = (\sigma, l, \sigma_{n})$, and estimate them using the training data. The parameters can be estimated by minimizing the negative marginal log-likelihood~\citep{GPframework,stein1988asymptotically,zhang2004inconsistent}:
\begin{align}
    -\log [p(\mathbf{Y}|\mathbf{X},\theta)] = \frac{1}{2} [(\mathbf{y} - \mathbf{\mu})^{\text{T}} K^{-1} (\mathbf{y} - \mathbf{\mu}) + \log |K| +N\log (2\pi)].
    \label{eqn: loglikelihood}
\end{align}
In the following section, we will show how the parameter updates are performed using the QHMC method. 
\subsection{Quantum-inspired Hamiltonian Monte Carlo}
QHMC is an enhanced version of the Hamiltonian Monte Carlo (HMC) algorithm that incorporates a random mass matrix for the particles, following a probability distribution. In conventional HMC, the position is represented by the original variables ($x$), while Gaussian momentum is represented by auxiliary variables ($q$). Utilizing the energy-time uncertainty relation of quantum mechanics, QHMC allows a particle to have a random mass matrix with a probability distribution. Consequently, in addition to the position and momentum variables, a mass variable ($m$) is introduced within the QHMC framework. Having a third variable offers the advantage of exploring various landscapes in the state-space. As a result, unlike standard HMC, QHMC can perform well on discontinuous, non-smooth and spiky distributions~\citep{barbu2020monte, Q-HMC}.

The quantum nature of QHMC can be understood by considering a one-dimensional harmonic oscillator example provided in~\citep{Q-HMC}. Let us consider a ball with a fixed mass $m$ attached to a spring at the origin. Assuming $x$ is the displacement, the magnitude of the restoring force that pulls back the ball to the origin is $F=-kx$, and the ball oscillates around the origin with period $T = 2\pi \sqrt{\frac{m}{k}}$. In contrast to standard HMC where the mass $m$ is fixed at 1, QHMC incorporates a time-varying mass, allowing the ball to experience acceleration and explore various distribution landscapes. That is, QHMC has the capability to employ a short time period $T$, corresponding to a small mass $m$, to efficiently explore broad but flat regions. Conversely, in spiky regions, it can switch to a larger time period $T$, \textit{i.e.} larger $m$, to ensure thorough exploration of all corners of the landscape~\citep{Q-HMC}. 

The implementation of QHMC is straightforward: we construct a stochastic process $M(t)$ for the mass, and at each time $t$, we sample $M(t)$ from a distribution $P_{M}(M)$. Resampling the positive-definite mass matrix is the only additional step to the standard HMC procedure. In practice, assuming that $P_{M}(M)$ is independent of $x$ and $q$, a mass density function $P_{M}(M)$ with mean $\mu_{m}$ and variance $\sigma^{2}_{m}$ can be $ \log m \sim \mathcal{N}(\mu_{m},\sigma^{2}_{m}),~ M = mI,$
where $I$ is the identity matrix. QHMC framework simulates the following dynamical system: 
\begin{align*}
    d \begin{pmatrix}
     x\\
     q
\end{pmatrix}= dt \begin{pmatrix}
     M(t)^{-1}q\\
     -\nabla U(x)
     \end{pmatrix}.
\end{align*}
 In this setting, the potential energy function of the QHMC system is $U(x) = -\log [p(\mathbf{Y}|\mathbf{X},\theta)]$, i.e., the negative of marginal log-likelihood. We summarize the algorithm in Algorithm~\ref{alg:QHMC}, and, here, we consider the location variables $\{x^{(i)}\}_{i=1}^{N}$ in GP model as the position variables $x$ in Algorithm~\ref{alg:QHMC}. The method evolves the QHMC dynamics to update the locations $x$.
 In this work, we implement the QHMC method for inequality constrained GP regression in a probabilistic manner.
\subsection{Proposed method}
Instead of enforcing all constraints strictly, the approach introduced in \citep{nonnegativityXiu} minimizes the negative marginal log-likelihood function in Equation~\ref{eqn: loglikelihood} while allowing constraint violations with a small probability. For example, for non-negativity constraints, the following requirement is imposed to the problem:
\begin{align*}
    P[(\mathbf{Y}(x)|x,\theta)<0] \leq \eta, \quad 
\text{for all} \quad x \in \mathcal{D},
\end{align*}
where $0< \eta <<1$.

In contrast to enforcing the constraint via truncated Gaussian assumption~\citep{maatouk2017gaussian} or performing inference based on the Laplace approximation and expectation propagation~\citep{jensen2013bounded}, the proposed method preserves the Gaussian posterior of the standard GP regression. The method uses a slight modification of the existing cost function. Given a model that follows a Gaussian distribution, we can re-express the constraint by the posterior mean and posterior standard deviation:
\begin{align}
    y^*(x) + \phi^{-1}(\eta)s(x) \geq 0, \quad \text{for all} \quad x \in \mathcal{D},
\end{align}
where $y^*(x)$ stands for the posterior mean, $s$ is the standard deviation and $\phi$ is the cumulative distribution function of a Gaussian random variable~\citep{nonnegativityXiu}. A practical choice for $\eta$ might be $\eta = 2.2\%$, resulting in $\phi^{-1}(\eta) = -2$. Then, we can formulate the optimization problem as 
\begin{align}
    \text{argmin}_{\theta} \quad &-\log [p(\mathbf{Y}|\mathbf{X},
    \theta)] \quad \text{such that} \label{eqn: first} \\ 
    &y^*(x) - 2s(x) \geq 0.
    \label{eqn: constraint}
\end{align}
In this particular form of the optimization problem, we encounter a functional constraint described by~\ref{eqn: constraint}. It can be prohibitive or impossible to satisfy this constraint at all points across the entire domain. Therefore, we adopt a strategy where we enforce Equation~\ref{eqn: constraint} only on a selected set of $m$ constraint points denoted as $ \mathbf{X}_{c} = {x_{c}^{(i)}}_{i=1}^{m}$. We reformulate the optimization problem as
\begin{align}
    \text{argmin}_{\theta} \quad &-\log [p(\mathbf{Y}|\mathbf{X},
    \theta)] \quad \text{such that}  \label{eqn: final1} \\ 
    &y^*(x_{c}^{(i)}) - 2s(x_{c}^{(i)}) \geq 0 \quad \text{for all} \quad i = 1,2,...,m, 
    \label{eqn: finalform}
\end{align}
where we estimate hyperparameters to enforce bounds. 
Solving this optimization problem can be very challenging, and hence, in~\citep{nonnegativityXiu} additional regularization terms are added.
Rather than directly solving the optimization problem, we adopt the soft-QHMC method, which introduces inequality constraints with a high probability (e.g., 95\%) by selecting a specific set of constraint points in the domain. We then enforce non-negativity on the posterior Gaussian Process at these selected points.
We minimize the log-likelihood in Equation~\ref{eqn: loglikelihood}, using the QHMC algorithm. Leveraging the Bayesian estimation~\citep{BayesBook}, we can approximate the posterior distribution by log-likelihood function and prior probability distribution as shown in the following:
\begin{align*}
     p(\mathbf{X}, \theta| \mathbf{Y}) \propto p(\mathbf{X},\theta, \mathbf{Y}) 
     = p(\theta)p(\mathbf{X}|\theta)p(\mathbf{Y}|\mathbf{X},\theta).
\end{align*}
The QHMC training flow starts with this Bayesian learning and proceeds with an MCMC procedure for drawing samples generated by the Bayesian framework. A general sampling procedure at step $t$ is given as 
\begin{align*}
    X^{(t+1)} &\sim \pi(X|\theta) = p(X|\theta^{(t)},Y), \\
    \theta^{(t+1)} &\sim \pi(\theta|X) = p(\theta|X^{(t+1)},Y).
\end{align*}

\begin{algorithm}[thb]
\textbf{Input:} Initial point $x_{0}$, step size $\epsilon$, number of simulation steps $L$, mass distribution parameters $\mu_{m}$ and $\sigma_{m}$.
\begin{algorithmic}[thbp]
\FOR{$t=1,2,...$} \STATE
 Resample $M_{t} \sim P_{M}(M)$\newline
 Resample  $q_{t} \sim N(0,M_{t})$\newline
 $(x_0,q_0) = (x^{(t)},q^{(t)})$ \newline
 $q_0 \leftarrow q_0 -\frac{\epsilon}{2}\nabla U(x_0)$
\FOR{$i=1,2,...,L-1$} \STATE
 $x_i \leftarrow x_{i-1} + \epsilon M_{t}^{-1}q_{i-1}$\newline
$q_i \leftarrow q_{i-1}-\frac{\epsilon}{2}\nabla U(x_i) $
\ENDFOR \newline
$x_L \leftarrow x_{L-1} + \epsilon M_{t}^{-1}q_{L-1}$\newline
$q_L \leftarrow q_{L-1}-\frac{\epsilon}{2}\nabla U(x_L)$\newline
$(\hat{x},\hat{q}) = (x_L,q_L)$ \newline
\textbf{MH step: } $u\sim$ \text{Uniform}$[0,1]$; \newline
$\rho = e^{-H(\hat{x},\hat{q})+H(x^{(t)},q^{(t)})} $;
\IF{$u<\min(1,\rho)$} 
\STATE $(x^{(t+1)},q^{(t+1)}) = (\hat{x},\hat{q})$ 
\ELSE 
\STATE $(x^{(t+1)},q^{(t+1)} = (x^{(t)},q^{(t)})$
\ENDIF
\ENDFOR \newline
\textbf{Output:} $\{x^{(1)},x^{(2)},...\} $
\end{algorithmic}
\caption{QHMC Training for GP with Inequality Constraints}
\label{alg:QHMC}
\end{algorithm}

\subsubsection{Enforcing Monotonicity Constraints}
We enforce the monotonicity constraints on a GP using the likelihood of derivative observations. We select active constraints and enforce the non-negativity on the partial derivative, \textit{i.e.}
\begin{align}
    \frac{\partial f}{\partial x_{i}}(\mathbf{x_{i}}) \geq 0, 
\label{eqn: monotonnonneg}
\end{align}
where $f$ is a vector of $N$ latent values. 
In our soft-constrained method, we introduce the non-negativity information in~\ref{eqn: monotonnonneg} on a set of selected points, and apply the same procedure as in~\ref{eqn: final1}. Since the derivative is also a GP with with mean and covariance matrix~\citep{riihimaki2010derivative}:

\begin{equation*}
    \mu(x') = \mathbb{E} \left [ \frac{\partial Y(x)}{\partial x_{i}} \right ], \quad \text{and} \quad 
    K(x,x') = \frac{\partial }{\partial x_{i}} \frac{\partial }{\partial x'_{i}} K(x,x'),
\end{equation*}
we can state the new posterior distribution as
\begin{align*}
    & p(\mathbf{y}^{*}, \theta| \mathbf{y}, \mathbf{x}, \mathbf{x}^{*}) = \int  p(\mathbf{y}^{*}, \theta| f^{*}) p(f^{*}| \mathbf{y}, \mathbf{x}, \mathbf{x}^{*}) df, \\
  &   p(f^{*}| \mathbf{y}, \mathbf{x}, \mathbf{x}^{*}) = \int 
     \int  p(f^{*}| \mathbf{x}^{*}, f, f')p(f,f'|\mathbf{x},\mathbf{y}) df df',
\end{align*}
where $\mathbf{y}^{*}$ and $\mathbf{f}^{*}$ denote the predictions at location $\mathbf{x}^{*}$.

\section{Theoretical analysis of the method}
In this section, using Bayes' Theorem, we will first show that QHMC can generate a steady-state distribution that approaches the true posterior distribution. Then, we present the convergence properties of the probabilistic approach on the optimization problem in the form of~\ref{eqn: finalform}.
\subsection{Convergence of QHMC training}
The study presented in~\citep{Q-HMC} demonstrates that the QHMC framework can effectively capture a correct steady-state distribution that describes the desired posterior distribution $p(x)\propto \exp(-U(x))$ via Bayes' rule.
The joint probability density of $(x,q,M)$ can be calculated by Bayesian theorem:
\begin{align*}
    p(x,q,M) = p(x,q|M)P_{M}(M),
\end{align*} 
where the conditional distribution is approximated as follows:
\begin{align*}
    p(x,q|M) \propto \exp{\left(-U(x)-K(q)\right)} = \exp{(-U(x))}\exp{\left(-\frac{1}{2}q^{T}M^{-1}q\right)}.
\end{align*}
Then, we can obtain the following 
\begin{align*}
    p(x) = \int_{q}\int_{M} dq dM p(x,q,M) \propto \exp(-U(x)),
\end{align*}
which shows that the marginal steady distribution approaches the true posterior distribution~\citep{Q-HMC}.
\subsection{Convergence properties of probabilistic approach}
\label{section: theorem}
In this section, we show that satisfying the constraints on a set of locations $x$ in the domain $\mathcal{D}$ will preserve convergence. Recall that we solve the following inequality-constrained optimization problem:
\begin{align*}
    \text{argmin}_{\theta} \quad &-\log [p(\mathbf{Y}|\mathbf{X},
    \theta)] \quad \text{such that}  \\ 
    &y^*(x_{c}^{(i)}) - 2s(x_{c}^{(i)}) \geq 0 \quad \text{for all} \quad i = 1,2,...,m.    
\end{align*}

Now, we need to show that the result obtained by using selected set of input locations will converge to the value of the regression model's output. This convergence ensures that probabilistic approach will eventually result in a model that satisfy the desired conditions. 

We use the assumption of $\mathcal{D}$ being finite throughout the proof. The proof can be constructed for the cases whether the domain is countable or uncountable.

(i) Assume that the domain $\mathcal{D}$ is a countable set containing $N$ elements. We select a subset $\mathcal{D}_{m} \in \mathcal{D}$ with $m$ points, where $x_{c}^{(1)},x_{c}^{(2)},...,x_{c}^{(m)} \in \mathcal{D}_{m}$. Each point $x \in \mathcal{D}$ has an associated Gaussian probability distribution, and we define the set of distributions of $x\in \mathcal{D}$ as $\mathcal{P}$. For the constraint points $x\in \mathcal{D}_{m}$, we have $m$ constraints and their corresponding probability distributions, which we define as $\mathcal{P}_{m}$. Additionally, we introduce a set $H(x)$ such that
\begin{align*}
    H(x):= \{ \theta| p(\mathbf{Y}|\mathbf{X}, \theta) < 0\},
\end{align*} 
which covers the locations where the non-negativity constraint is violated. For each fixed $x\in \mathcal{D}$, we define 
\begin{align*}
    v(x) &:= \inf_{P \in \mathcal{P}} P(\mathbf{Y}|\mathbf{X}, \theta) < 0 \equiv \inf_{P \in \mathcal{P}} P(H(x)), \quad \text{and} \\
    v_{m}(x) &:= \inf_{P \in \mathcal{P}_{m}} P(\mathbf{Y}|\mathbf{X}, \theta) < 0 \equiv \inf_{P \in \mathcal{P}_{m}} P(H(x)).
\end{align*}
(ii) Assume that the domain $\mathcal{D}$ is a finite but uncountable set. In this case, we can construct a countable subset $\mathcal{\Tilde{D}}$ such that $x\in \mathcal{\Tilde{D}}$. We define the set of probability distributions as in case (i). Since $\mathcal{D}$ is finite, the set $\mathcal{D} \cup \{x\} $ is also finite. Consequently, the sets $H(x), v(x)$ and $v_{m}(x)$ can be constructed as in the first case. Now, we will establish a convergence of $v_{m}$ over $v$ as $\mathcal{P}_{m}$ converges to $\mathcal{P}$.

First, let us provide distance metrics used throughout the proof. Following the definitions in~\citep{proofChance}, let 
\begin{align*}
    d(x,A):= \inf_{x'\in A} ||x-x'||
\end{align*}
denote the distance from a point $x$ to a set $A$. Using this, we can define the distance of two compact sets $A$ and $B$ as \begin{align*}
    \mathbb{D}(A,B) := \sup_{x\in A} d(x,B).
\end{align*}  
Then, the Hausdorff distance between $A$ and $B$ is defined as $\mathbb{H}(A,B) := \max\{\mathbb{D}(A,B), \mathbb{D}(B,A)\}$. Finally, we define a pseudo-metric $\mathbf{d}$ to describe the distance between two probability distributions $P$ and $\Tilde{P}$ as
\begin{align*}
    \mathbf{d}(P,\Tilde{P}) := \sup_{x \in \mathcal{D}}|P(H(x))-\Tilde{P}(H(x))|,
\end{align*}
where $\mathcal{D}$ is the domain specified in Section~\ref{section: theorem}. 
\begin{assumption}
We assume that the probability distributions $\mathcal{P}$ and $\mathcal{P}_{m}$ satisfy the following conditions:
\begin{enumerate}
    \item There exists a weakly compact set $\mathcal{\Tilde{P}}$ such that $\mathcal{P} \subset \mathcal{\Tilde{P}}$ and $\mathcal{P}_{m} \subset \mathcal{\Tilde{P}}$.
    \item $\underset{m\rightarrow N}{\lim} \mathbf{d}(\mathcal{P}, \mathcal{P}_{m}) = 0,$ with probability 1.
    \item $\underset{m\rightarrow N}{\lim} \mathbf{d}(\mathcal{P}_{m}, \mathcal{P}) = 0$, with probability 1.
\end{enumerate}
\label{assumption}
\end{assumption}
Now, we show that Theorem~\ref{thm: convergence} holds under the assumptions in Assumption~\ref{assumption}. Recall that we have
\begin{align*}
        \mathbb{H}(\text{conv}V,\text{conv}V_{m}) = \max \left \{ \left | \sup_{P \in \mathcal{P}_{m}}P(H(x)) - \sup_{P \in  \mathcal{P}} P(H(x)) \right| , \left | \inf_{P \in  \mathcal{P}_{m}}P(H(x)) - \inf_{P \in  \mathcal{P}} P(H(x)) \right| \right \}.
    \end{align*}
Based on the definition and property of Hausdorff distance~\citep{Hausdorff} we also have
   \begin{align*}
        \mathbb{H}(\text{conv}V,\text{conv}V_{m}) \leq \mathbb{H}(V, V_{m}) \leq \max\{\mathbb{D}(V,V_{m}), \mathbb{D}(V_{m},V)  \}.
    \end{align*}
Here we have
\begin{align*}
    \mathbb{D}(V,V_{m}) &= \sup_{v\in V} \inf_{v'\in V_{m}} ||v-v'|| \\
     & = \sup_{P \in \mathcal{P}} \inf_{\Tilde{P} \in \mathcal{P}_{m}} ||P(H(x))-\Tilde{P}(H(x))|| \\
     & \leq \sup_{P \in \mathcal{P}} \inf_{\Tilde{P} \in \mathcal{P}_{m}} \sup_{x\in \mathcal{D}} ||P(H(x))-\Tilde{P}(H(x))|| \\
     &= \mathbf{d}(\mathcal{P}, \mathcal{P}_{m}).
\end{align*}
Applying the same procedure, we also obtain $\mathbb{D}(V_{m},V) \leq \mathbf{d}(\mathcal{P}_{m}, \mathcal{P})$. Hence, 
\begin{align*}
        \mathbb{H}(\text{conv}{V}, \text{conv}{V_{m}}) \leq \mathbb{H} (V,V_{m}) \leq \mathbb{H}(\mathcal{P}_{m}, \mathcal{P}).
\end{align*}
Consequently, we obtain
\begin{align*}
        |v_{m}(x) - v(x)|& \leq  \left | \inf_{P\in \mathcal{P}_{m}} P(H(x)) - \inf_{P\in \mathcal{P}} P(H(x)) \right | \\ 
        &\leq \mathbb{H}(\text{conv}{V}, \text{conv}{V_{m}}) \\
        &\leq \mathbb{H}( \mathcal{P}_{m},  \mathcal{P}).
\end{align*}
\begin{theorem}
$v_{m}$ converges to $v$ as $\mathcal{P}_{m}$ converges to $\mathcal{P}$, that is
\begin{align*}
    \lim_{m \rightarrow N} \sup_{x \in \mathcal{D}} |v_{m}(x) - v(x)| = 0.
\end{align*}
    \begin{proof}
     Let us assume that $x \in \mathcal{D}$ is fixed, and define
     \begin{align*}
         V := \{P(H(x)) : P \in \text{cl}\mathcal{P} \}, \quad \text{and}, \quad V_{m} := \{P(H(x)) : P \in \text{cl}\mathcal{P}_{m} \},
     \end{align*}
     where cl represents the closure. Note that both $V$ and $V_{m}$ are bounded subsets in $\mathbb{R}^{d}$. Let us define $a,b,a_{m}$ and $b_{m}$ such that
     \begin{align*}
         a := \inf_{v\in V}  v, \quad b := \sup_{v\in V}  v, \quad  a_{m} := \inf_{v\in V_{m}} v, \quad b_{m} := \sup_{v\in V_{m}}  v, 
     \end{align*}
     The Hausdorff distance between convex hulls~(conv) of the sets $V$ and $V_{m}$ are calculated as~\citep{Hausdorff}
     \begin{align*}
          \mathbb{H}(\text{conv}V,\text{conv}V_{m}) = \max \{ |b_{m}-b|, |a-a_{m}| \}.
     \end{align*}
    Since we know that
    \begin{align*}
        b_{m}-b = \sup_{v\in V_{m}} v - \sup_{v\in V}  v, \quad \text{and} \quad 
        a_{m} - a = \inf_{v\in V_{m}} v -  \inf_{v\in V}  v,
    \end{align*}
    we have 
    \begin{align*}
        \mathbb{H}(\text{conv}V,\text{conv}V_{m}) = \max \left \{ \left | \sup_{P \in \mathcal{P}_{m}}P(H(x)) - \sup_{P \in  \mathcal{P}} P(H(x)) \right| , \left | \inf_{P \in  \mathcal{P}_{m}}P(H(x)) - \inf_{P \in  \mathcal{P}} P(H(x)) \right| \right \}
    \end{align*}
    Based on the definition and property of Hausdorff distance~\citep{Hausdorff} we have
    \begin{align*}
        \mathbb{H}(\text{conv}V,\text{conv}V_{m}) \leq \mathbb{H}(V, V_{m}),
    \end{align*}
    resulting in~\citep{proofChance}
    \begin{align*}
        |v_{m}(x) - v(x)| \leq \mathbb{H}(V, V_{m}) \leq \mathbb{H}( \mathcal{P},  \mathcal{P}_{m}).
    \end{align*}
    In this setting, $x$ can be any point in $\mathcal{D}$, and the right hand side of the inequality is independent of $x$. We can complete the proof by taking the supremum of each side with respect to $x$~\citep{proofChance}.
     \end{proof}
     \label{thm: convergence}
\end{theorem} 
\section{Numerical examples}
\label{section: results}
In this section, we evaluate the performance of the proposed algorithms on various examples including synthetic and real data. We conduct experiments on datasets that have different scales of sizes and dimensions. We introduce several versions of QHMC algorithms depending on the selection of constraint point locations and probabilistic approach. 

Rather than randomly locating $m$ constraint points, we start with an empty constraint set and determine the locations of the constraint points one by one adaptively. Throughout this process, we employ various strategies for adding the constraints. The specific approaches are outlined as follows:
\begin{enumerate}
    \item Constraint-adaptive approach: This approach examines whether the constraint is satisfied at a location. The function value is calculated, and if the constraint is violated at that location, then a constraint point is added. 
    \item Variance-adaptive approach: We calculate the prediction variance in the test set. We select constraint points at the locations where we observe the largest variance values. The goal is basically to reduce the variance in predictions and increase the stability.
    \item Combination of constraint and variance adaption: In this approach, we determine a threshold value~(e.g. $0.20$) for the variance, and the algorithm locates constraint points to the locations where the largest prediction variance is observed. Once the variance reduces to the threshold value, the algorithm switch to the first strategy, in which it locates constraint points where the violation occurs. 
\end{enumerate}

We represent the constraint-adaptive, hard-constrained approach as QHMCad and its soft-constrained counterpart as QHMCsoftad. Similarly, QHMCvar refers to the method focusing on variance, while QHMCsoftvar corresponds to its soft-constrained version.
We denote the combination of the two approaches with hard and soft constraints by QHMCboth and QHMCsoftboth, respectively. For the sake of comparison, we implement the truncated Gaussian algorithms using an HMC sampler~(tnHMC) and a QHMC sampler~(tnQHMC) for inequality-constrained examples, while we implement additive GP~(additiveGP) algorithm for monotonicity-constrained examples. 

For the synthetic examples, we evaluated the time and accuracy performances of the algorithms while simultaneously changing the dataset size and noise level in the data. Following~\citep{nonnegativityXiu}, as our metric, we calculate the relative $l_{2}$ error between the posterior mean $y^{*}$ and the true value of the target function $f(x)$ on a set of test points $\mathbf{X}_{t} = \{x_{T}^{(i)}\}_{i=1}^{N_{t}}$:
\begin{align*}
    E = \sqrt{\frac{\sum_{i=1}^{N_{t}}[y^{*}(x_{T}^{(i)})-f(x_{T}^{(i)})]^{2}}{\sum_{i=1}^{N_{t}}f(x_{T}^{(i)})^{2}}}.
\end{align*}
Additionally, in order to highlight the advantage of QHMC over HMC, we implement our approach using the standard HMC sampling procedure. The relative error, posterior variance and execution time of each version of QHMC and HMC algorithms are presented. 
\subsection{Inequaltiy Constraints}
In this section, we provide two synthetic examples and two real-life application examples to demonstrate the effectiveness of our algorithms on inequality constraints. In synthetic examples, we compare the performance our approach with truncated Gaussian methods for a 2-dimensional and a 10-dimensional problems.
For the 2-dimensional example, our primary focus is on enforcing the non-negativity constraints within the GP model. In the case of the 10-dimensional example, we generalize our analysis to satisfy a different inequality constraint. We evaluate the performances of truncated Gaussian, QHMC and  soft-QHMC methods. In third example, we consider conservative transport in a steady-state velocity field in heterogeneous porous media. Despite being a two-dimensional problem, the non-homogeneous structure of the solute concentration introduces complexity and increases the level of difficulty. The last example is a 3-dimensional heat transfer problem in a hallow sphere. 
\subsubsection{Example 1}
We consider the following 2D function under non-negativity constraints:
\begin{align*}
    f(x) = \arctan{5x_{1}} + \arctan{x_{2}},
\end{align*}
where $\{x_{1},x_{2}\}\in [0,1]^{2}$. We train our GP model via QHMC over $20$ randomly selected locations. 

\begin{figure}[ht]
\begin{center}
\includegraphics[width=15cm]{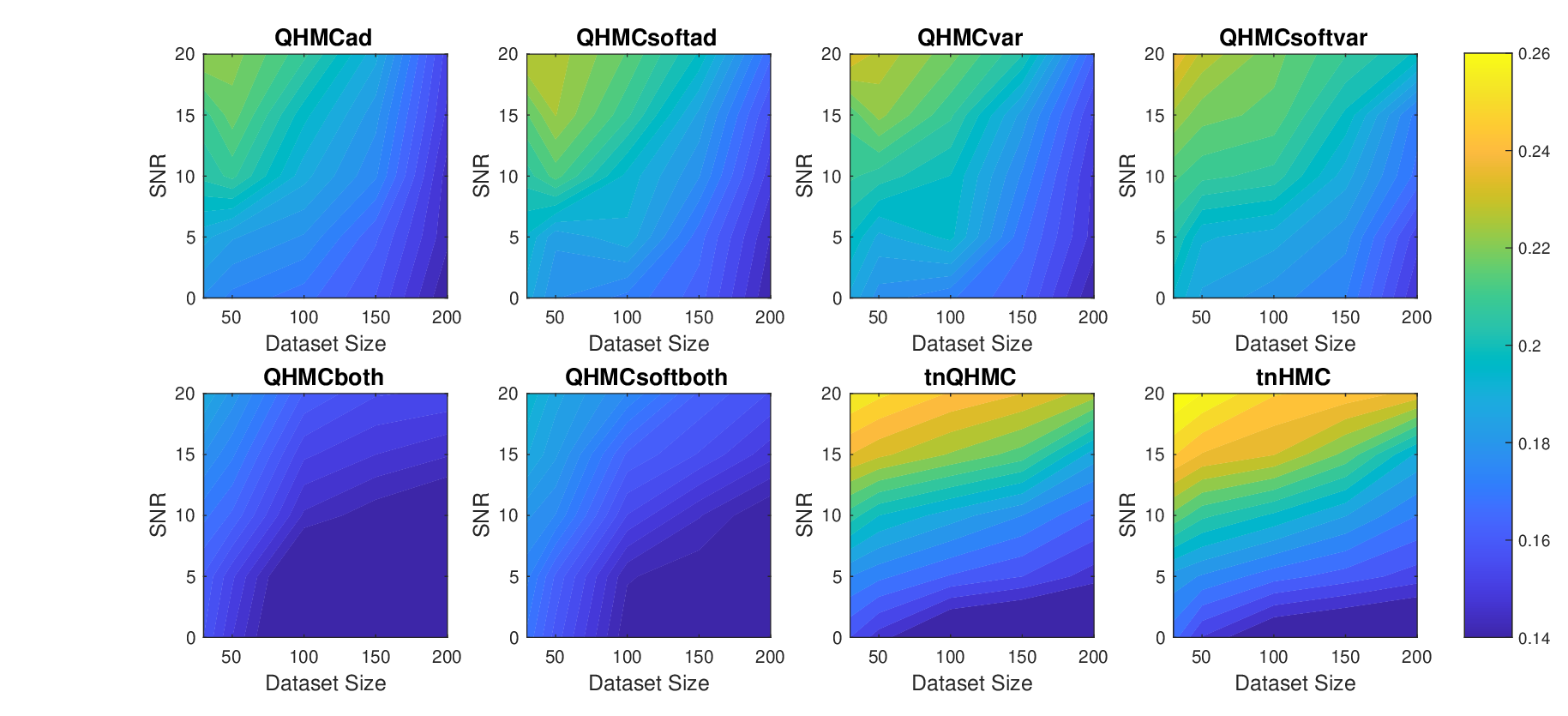}
\end{center}
\caption{Relative error of the algorithms with different data sizes and signal to noise ratios~(SNR) for Example 1~(2D), inequality.}
\label{fig: 2Derror}
\end{figure}

Figure~\ref{fig: 2Derror} presents the relative error values of the algorithms with respect to two parameters: the size of the dataset and signal-to-noise ratio~(SNR). 
It can be seen that the most accurate results without adding any noise are provided by QHMCboth and tnQHMC algorithms with around $10\%$ relative error. 
However, upon introducing the noise to the data and increasing its magnitude, we observe a distinct pattern. The QHMC methods exhibit relative error values of approximately $15\%$ within the SNR range of $15\%$ to $20\%$. In contrast, the relative error of the truncated Gaussian methods increases to $25\%$ within the same noise range. This pattern demonstrates that QHMC methods can tolerate noise and maintain higher accuracy under these conditions. 

In Table~\ref{table: QHMCvsHMC2Dineq}, the comparison between QHMC and HMC algorithms with a dataset size of 200 is presented. The relative error values indicate that QHMC yields approximately $20\%$ more accurate results than HMC, and it achieves this with a shorter processing time. Consequently, QHMC demonstrates both higher accuracy and efficiency compared to HMC.

\begin{table}
\begin{center}
\caption[Comparison of QHMC and HMC on 2D, inequality.]{Comparison of QHMC and HMC on 2D, inequality.}
 \begin{tabular}{ |l || c | c| c| l ||c | c | c| c| r }
\hline

\hline 
 Method &  Error &  Posterior Var & Time  & Method &  Error &  Posterior Var & Time  \\ \hline \hline
\multirow{1}{*}{QHMC-ad} & 0.10 & 0.14 & 46s & {HMC-ad}& 0.12 & 0.17 & 52s \\
\hline
\multirow{1}{*}{QHMC-soft-ad} & 0.11  & 0.16  & 39s &{HMC-soft-ad} & 0.13 & 0.19 & 48s \\
\hline

\multirow{1}{*}{QHMC-var}  &  0.11 & 0.12  & 40s & {HMC-var} & 0.13 & 0.14 & 46s \\ \hline
\multirow{1}{*}{QHMC-soft-var}  & 0.12 & 0.15 & 34s & {HMC-soft-var} & 0.15 & 0.14 & 42s \\ \hline
\multirow{1}{*}{\textbf{QHMC-both}}  & \textbf{0.08} & 0.13 & 48s  & {HMC-both}& 0.10 & 0.14 & 53s \\ \hline
\multirow{1}{*}{\textbf{QHMC-soft-both}}  & 0.09 & 0.13 & 39s & {HMC-soft-both} & 0.12 & 0.15 & 44s\\ \hline
\end{tabular}
\label{table: QHMCvsHMC2Dineq}
\end{center}
\end{table}

\begin{figure}[ht]
\begin{center}
\includegraphics[width=15cm]{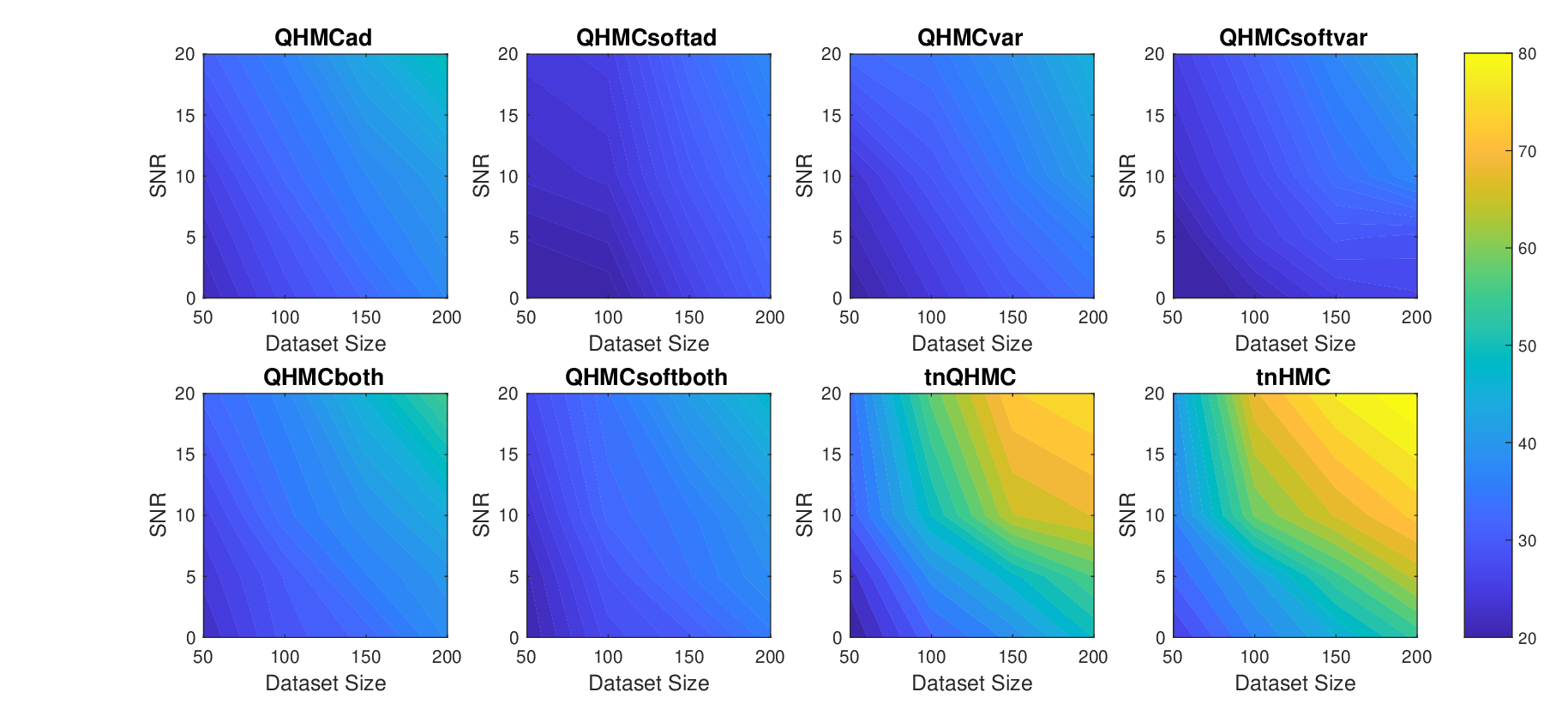}
\end{center}
\caption{Execution times~(in seconds) of the algorithms with different signal to noise ratios~(SNR) and datasizes for Example 1~(2D), inequality.}
\label{fig: coloredTime2D}
\end{figure}

\begin{figure}[ht]
\begin{center}
\includegraphics[width=15cm]{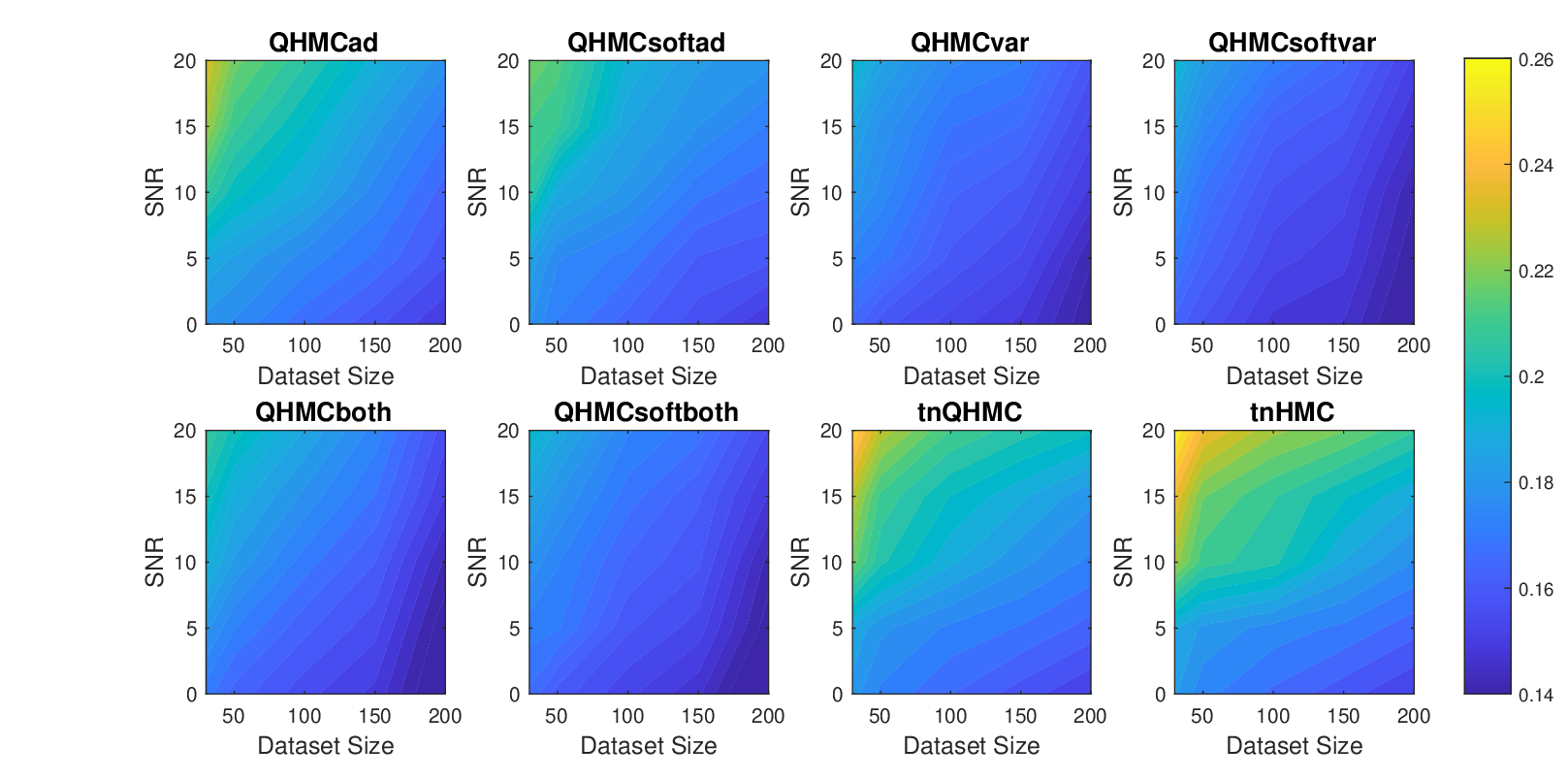}
\end{center}
\caption{Posterior variances of the algorithms with different signal to noise ratios~(SNR) and datasizes for Example 1~(2D), inequality.}
\label{fig: 2DcoloredVar}
\end{figure}

Further, we compare the time performances of the algorithms in Figure~\ref{fig: coloredTime2D} which demonstrates that QHMC methods, especially the probabilistic QHMC approaches can perform much faster than the truncated Gaussian methods. In this simple 2D example, the presence of noise does not significantly impact the running times of the QHMC algorithms. In contrast, truncated Gaussian algorithms are slower under noisy environment even when the dataset size is small. We can also observe in Figure~\ref{fig: 2DcoloredVar} that the QHMC algorithms, especially QHMCvar and QHMCboth are the most robust ones, as their small relative error comes with a small posterior variance. In contrast, the posterior variance values of the truncated Gaussian methods are higher than QHMC posterior variances even when there is no noise, and gets higher along with the relative error~(see Figure~\ref{fig: 2Derror}) when the SNR levels increase. Combining all of these experiments, we can conclude that QHMC methods achieve higher accuracy within a shorter time frame. Consequently, these methods prove to be more efficient and robust as they can effectively tolerate changes in parameters. 
Additionally, it is worth noting that we observed a slight improvement in the performance of truncated Gaussian algorithms by implementing tnQHMC. Based on the numerical results obtained by tnQHMC, it can be concluded that employing tnQHMC not only yields higher accuracy but also saves some computational time compared to tnHMC.

\subsubsection{Example 2}
Next, we consider the 10D Ackley function~\citep{SCBO} defined as follows:
        \begin{align*}
            f(x) = -a \exp{\left(-b \sqrt{\frac{1}{d}\sum_{i=1}^d x_{i}^2}\right)} - \exp{\left(-b \sqrt{\frac{1}{d}\sum_{i=1}^d \cos{cx_{i}}}\right)} + a + \exp{1},
        \end{align*}
where $d=10$, $a=20$, $b=0.2$ and $c=2\pi$. 
We study the performance of the algorithms on the domain $[-10,10]^{10}$ while enforcing the function to be greater than $5$.

\begin{table}
\begin{center}
\caption[Comparison of QHMC and HMC on 10D, inequality.]{Comparison of QHMC and HMC on 10D, inequality.}
 \begin{tabular}{ |l || c | c| c| l ||c | c | c| c| r }
\hline

\hline 
 Method &  Error &  Posterior Var & Time  & Method &  Error &  Posterior Var & Time  \\ \hline \hline
\multirow{1}{*}{QHMC-ad} & 0.10 & 0.13 & 39m 17s & {HMC-ad}& 0.12 & 0.15 & 43m 33s \\
\hline
\multirow{1}{*}{QHMC-soft-ad} & 0.11  & 0.14  & 36m 21s &{HMC-soft-ad} & 0.13 & 0.15 & 41m 10s \\
\hline

\multirow{1}{*}{QHMC-var}  &  0.11 & 0.11  & 37m 4s & {HMC-var} & 0.13 & 0.12 & 41m 31s \\ \hline
\multirow{1}{*}{QHMC-soft-var}  & 0.12 & 0.11 & 34m 23s & {HMC-soft-var} & 0.14 & 0.12 & 37m 42s \\ \hline
\multirow{1}{*}{\textbf{QHMC-both}}  & \textbf{0.09} & 0.12 & 40m 8s  & {HMC-both}& 0.10 & 0.14 & 44m 23s \\ \hline
\multirow{1}{*}{\textbf{QHMC-soft-both}}  & 0.10 & 0.12 & 37m 53s & {HMC-soft-both} & 0.12 & 0.14 & 42m 5s \\ \hline
\end{tabular}
\label{table: QHMCvsHMC10Dineq}
\end{center}
\end{table}

\begin{figure}[ht]
\begin{center}
\includegraphics[width=15cm]{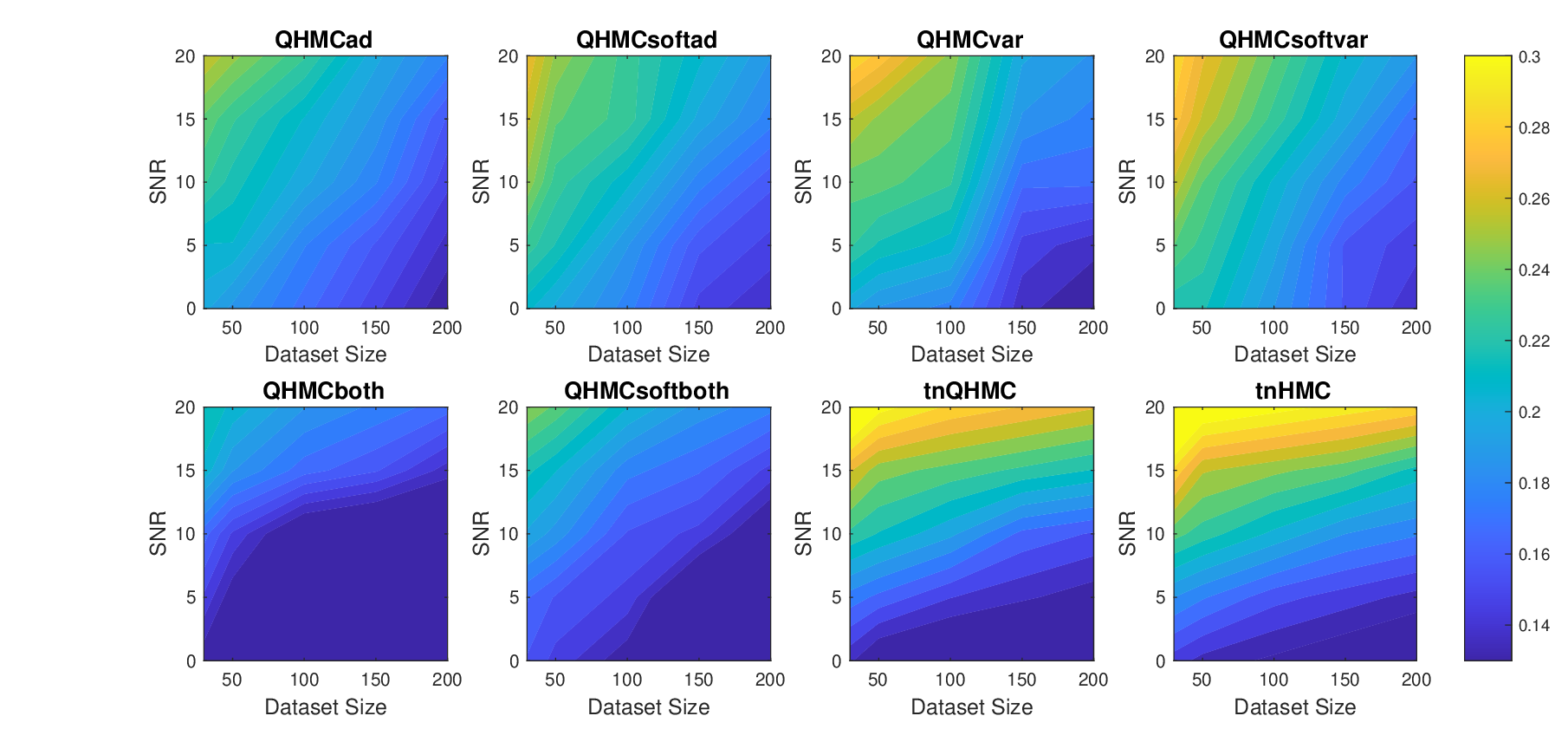}
\end{center}
\caption{Relative error of the algorithms with different data sizes and signal to noise ratios~(SNR) for Example 2~(10D), inequality.}
\label{fig: coloredrelativeError10d}
\end{figure}

Figure~\ref{fig: coloredrelativeError10d} illustrates that QHMCboth, QHMCsoftboth and truncated Gaussian algorithms yield the lowest error when there is no noise in the data. However, as the noise level increases, truncated Gaussian methods fall behind all QHMC approaches. 
Specifically, both the QHMCboth and QHMCsofthboth algorithms demonstrate the ability to tolerate noise levels up to $15\%$ with an associated relative error of approximately $15\%$. However, other variants of QHMC methods display greater noise tolerance when dealing with larger datasets. With fewer than $100$ data points, the error rate reaches around $25\%$, but it decreases to $15-20\%$ when the number of data points exceeds $100$. 

\begin{figure}[ht]
\begin{center}
\includegraphics[width=15cm]{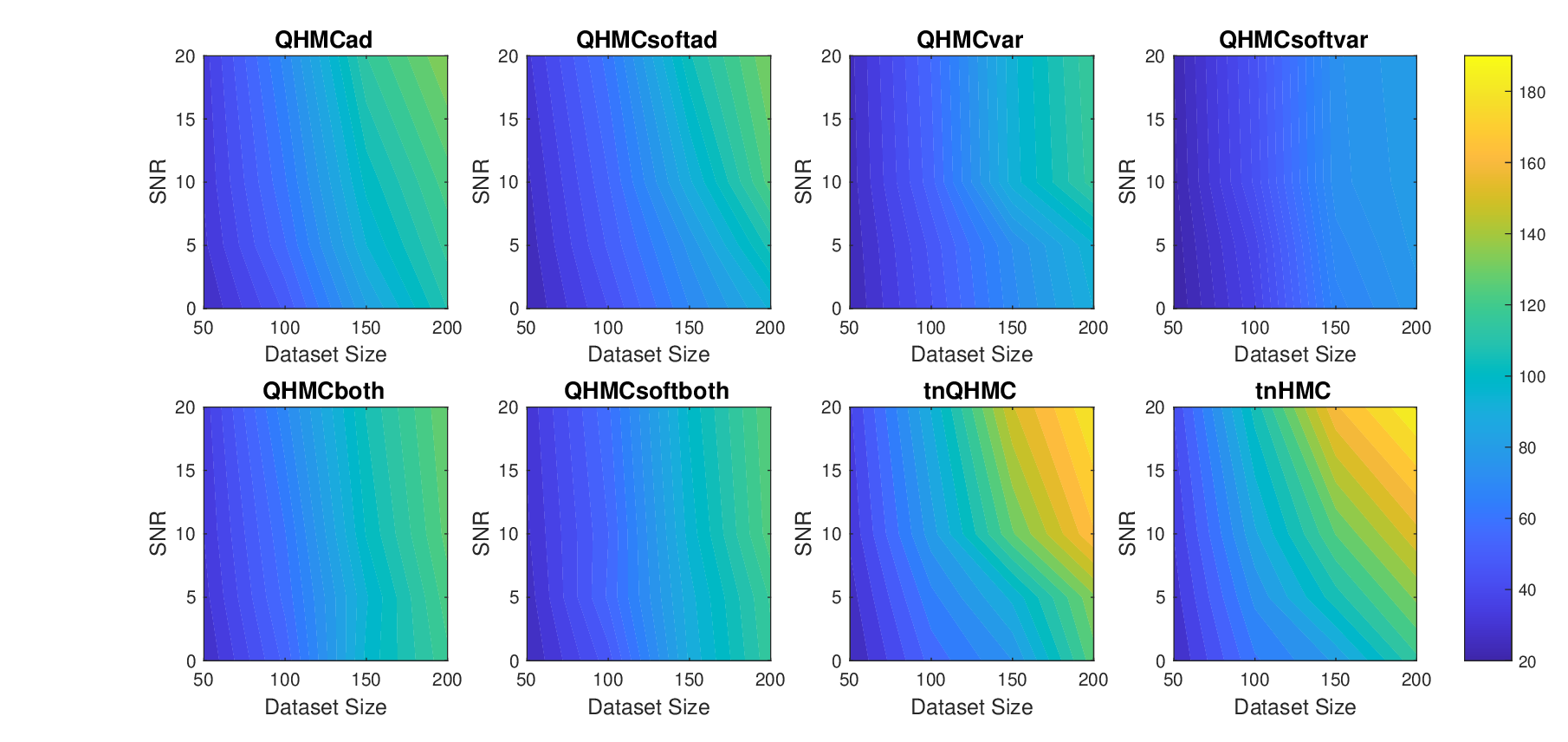}
\end{center}
\caption{Execution times~(in minutes) of the algorithms with different signal to noise ratios~(SNR) and datasizes for Example 2~(10D), inequality.}
\label{fig: coloredTime10D}
\end{figure}

Figure~\ref{fig: coloredTime10D} illustrates the time comparison of the algorithms, where we can observe that QHMC methods provide around $30-35\%$ time efficiency for the datasets larger than a size of $150$. Combining this time advantage with the higher accuracy of QHMC indicates that both soft and hard constrained QHMC algorithms outperform truncated Gaussian methods across various criteria. 
QHMC methods offer the flexibility to employ one of the algorithms depending on the priority of the experiments. For example, if speed is the primary consideration, QHMCsoftvar is the fastest method while maintaining a good level of accuracy. If accuracy is the most important metric, employing QHMCboth would be a wiser choice, as it still offers significant time savings compared to other methods.

\begin{figure}[ht]
\begin{center}
\includegraphics[width=15cm]{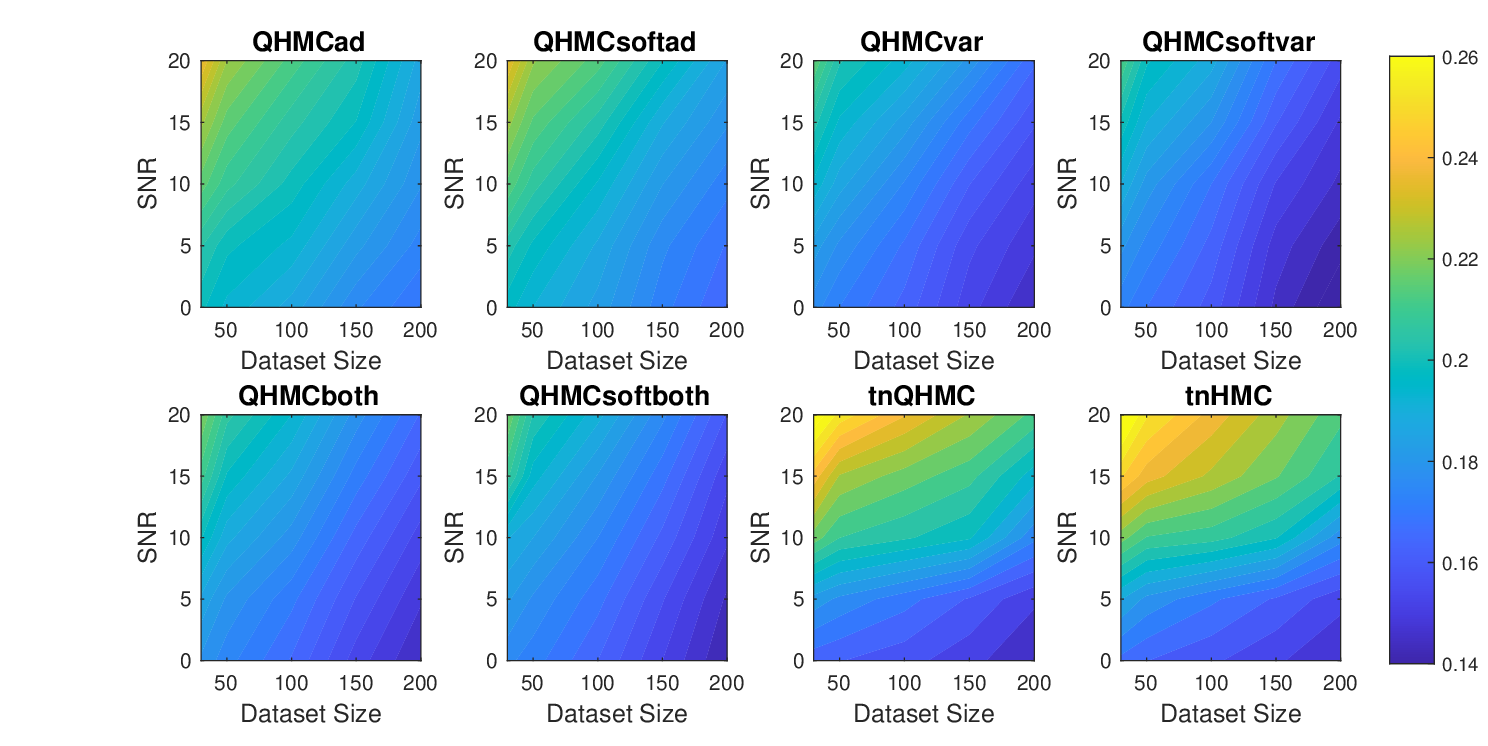}
\end{center}
\caption{Posterior variances of the algorithms with different signal to noise ratios~(SNR) and datasizes for Example 2~(10D), inequality.}
\label{fig: 10DcoloredVar}
\end{figure}

Figure~\ref{fig: 10DcoloredVar} presents that the posterior variance values of truncated Gaussian methods are significantly higher than that of the QHMC algorithms, especially when the noise levels are higher than $5\%$. As expected, QHMCvar and QHMCsoftvar algorithms offer the lowest variance, while QHMCboth and QHMCsoftboth follow them. A clear pattern is shown in the figure, in which QHMC approaches can tolerate higher noise levels especially when the dataset is large. 
It is notable that our method demonstrates a significant increase in efficiency as the dimension increases. When comparing this 10D example to the 2D case, the execution times of the truncated Gaussian methods are notably impacted by the dimension, even in the absence of noise in the datasets. Although their relative error levels can remain low without noise, it takes $1.5$ times longer than the QHMC methods to offer those accuracy. Additionally, this observation holds only for cases where the data is noise-free. As soon as noise is present, the accuracy of truncated Gaussian methods deteriorates, whereas QHMC methods can withstand the noise and yield good results in a shorter time span.

\subsubsection{Example 3: Solute transport in heterogeneous porous media}
\begin{figure}[ht]
\begin{center}
\includegraphics[width=12cm]{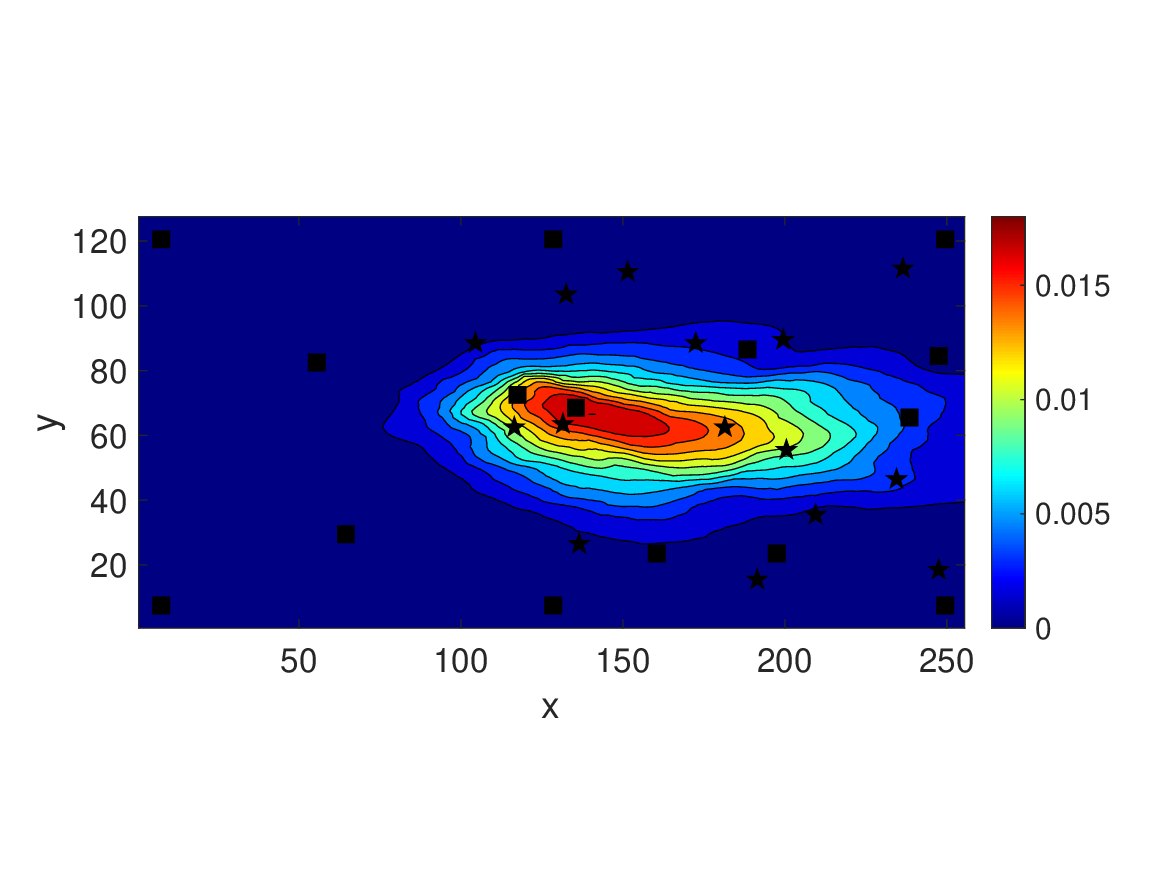}
\end{center}
\caption{Observation locations (black squares) and constraint locations (black stars).}
\label{fig: soluteMain}
\end{figure}

\begin{figure}[ht]
\begin{center}
\includegraphics[width=10cm]{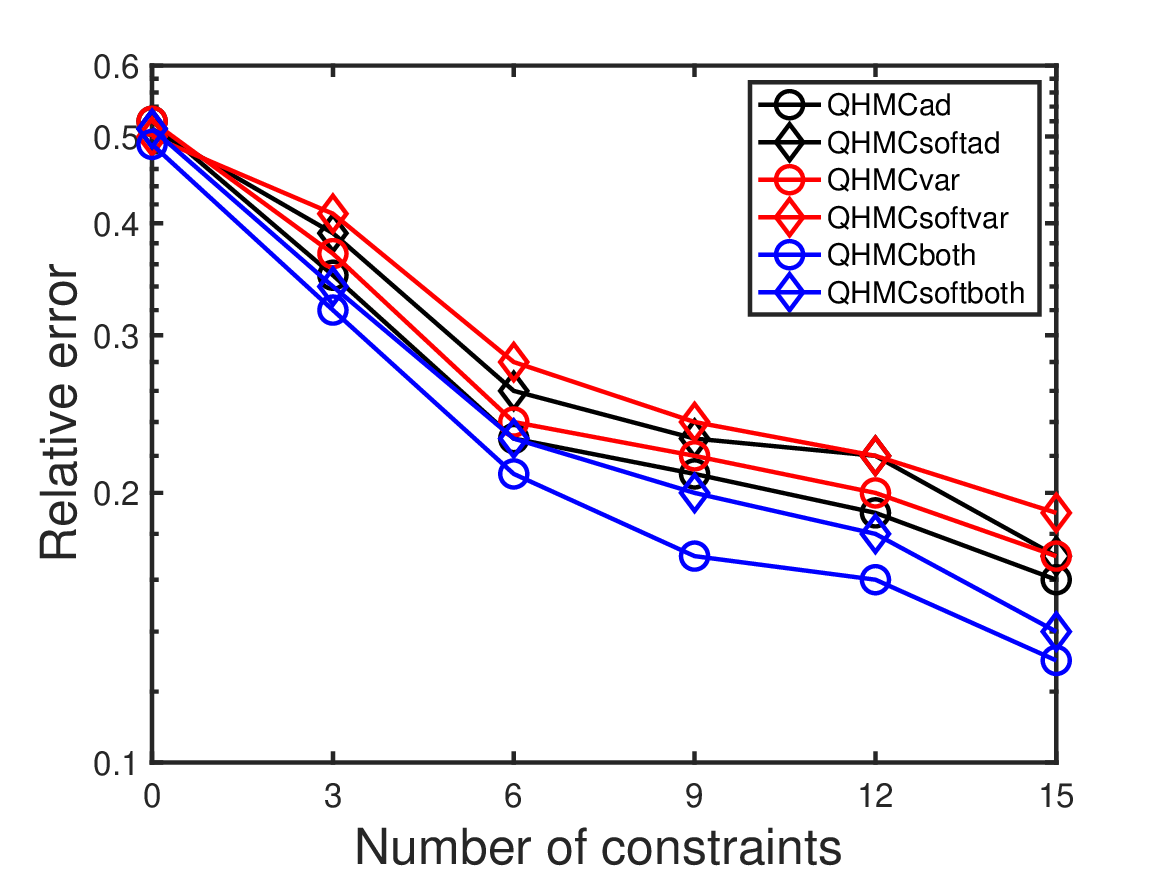}
\end{center}
\caption{The change in relative error while adding constraints, solute transport.}
\label{fig: numconstraintsSolute}
\end{figure}

Following the example in~\citep{yang2019physics}, we examine conservative transport within a constant velocity field in heterogeneous porous media. Let us denote the solute concentration by $C(\mathbf{x},t)(\mathbf{x} = (x,y)^\text{T})$, and suppose that the measurements of $C(\mathbf{x},t)$ are available at various locations at different times. Conservation laws can be used to describe the processes of flow and transport. Specifically, we can describe the flow using Darcy flow equation~\citep{yang2019physics}
\begin{align}
    \begin{cases}
    \nabla \cdot (K\nabla h) = 0, & \mathbf{x} \in \mathbb{D}, \\
    \frac{\partial h}{\partial \mathbf{n}} = 0, & y = 0 \text{ or } y = L_2, \\
    h = H_1, & x = 0, \\
    h = H_2, & x = L_1,
\end{cases}
\end{align}
where $h(\mathbf{x},w)$ is the hydraulic head, $\mathbb{D} = [0,L_{1}] \times [0,L_{2}]$ is the simulation domain with $L_{1} = 256$ and $L_{2} = 128$, $H_{1}$ and $H_{2}$ are known boundary head values and $K(\mathbf{x},w)$ is the unknown hydraulic conductivity field. The field is represented as a stochastic process, with the distribution of values described by a log-normal distribution. Specifically, it is expressed as $K(\mathbf{x},w) = \exp Z(\mathbf{x},w)$, where is a second-order stationary GP with a known exponential covariance function, $\text{Cov} \{Z(\mathbf{x}), Z(\mathbf{x}')\} = \sigma_{Z}^{2} \exp{(-|\mathbf{x} - \mathbf{x}'|/l_{z})}$ where variance $\sigma_{Z}^{2}=2$ and correlation length $l_{z}=5$. We can describe the solute transport by the advection-dispersion equation~\citep{emmanuel2005mixing, lin2009efficient,yang2019physics} as
\begin{align}
    \begin{cases}
    \frac{\partial C}{\partial t} + \nabla \cdot (v \mathbf{C}) = \nabla \cdot \left( \frac{D_{w}}{\tau} + \mathbf{\alpha} ||\mathbf{v}||_2 \right) \nabla C, & \mathbf{x} \text{ in } \mathbb{D}, \\
    C = Q \delta(\mathbf{x} - \mathbf{x}^*), & t = 0, \\
    \frac{\partial C}{\partial \mathbf{n}} = 0, & y = 0 \text{ or } y = L_2 \text{ or } x = L_1, \\
    C = 0, & x = 0.
    \end{cases}
\end{align}
In this context, $C(\mathbf{x},t;w)$ represents the solute concentration defined over $\mathbb{D} \times [0,T] \times \Omega $, $\mathbf{v}$ denotes the fluid velocity given by $\mathbf{v}(\mathbf{x};w) = -K(\mathbf{x};\omega)\nabla h(\mathbf{x},\omega)/\phi$ with $\phi$ being porosity; $D_{w}$ is the diffusion coefficient, $\tau$ stands for the tortuosity, and $\mathbf{\alpha}$  is the dispersivity tensor, with diagonal components $\alpha_{L}$ and $\alpha_{T}$. In this study, the transport parameters are defined as follows: $\phi=0.317, \tau = \phi^{1/3}, D_w = 2.5 \times 10^{-5},  \alpha_L = 5$ and $\alpha_T = 0.5$. Lastly, the solute is instantaneously injected at $\mathbf{x}^* = (50,64)$ at $t=0$ with the intensity $Q=1$~\citep{yang2019physics}. In Figure~\ref{fig: soluteMain}, the ground truth with observation locations and constraint locations are presented to provide an insight into the structure of solute concentration. 
Table~\ref{table: QHMCvsHMCsolute} presents a comparison of all versions of QHMC and HMC methods, along with the truncated Gaussian algorithms. Similar to the results observed with synthetic examples, the QHMCboth, QHMCsoftboth, and tnQHMC algorithms demonstrate the most accurate predictions with a relative error of $13-15\%$. Notably, QHMCsoftboth emerges as the fastest among the methods while achieving higher accuracy. For instance, the error value obtained by QHMCsoftboth is $0.14$, whereas tnQHMC's error is $0.15$. However, QHMCsoftboth delivers a $20\%$ time efficiency gain with slightly superior accuracy. In Figure~\ref{fig: numconstraintsSolute}, a comprehensive comparison of the algorithms is presented. We can observe the decrease in the relative error values while we add the constraints step by step, according the the adopted adaptive approach. Initially, the error is $0.5$ and gradually decreases to approximately $0.13$. Furthermore, it is evident that the QHMCboth and QHMCsoftboth methods consistently deliver the most accurate results at each step, whereas the performance of the QHMCsoftvar method is outperformed by other approaches.

\begin{table}
\begin{center}
\caption[Comparison of QHMC and HMC on solute transport with nonnegativity.]{Comparison of QHMC and HMC on solute transport with nonnegativity.}
 \begin{tabular}{ |l || c | c| c| l ||c | c | c| c| r }
\hline

\hline 
 Method &  Error &  Posterior Var & Time  & Method &  Error &  Posterior Var & Time  \\ \hline \hline
\multirow{1}{*}{QHMC-ad} & 0.18 & 0.13 & 83s & {HMC-ad}& 0.20 & 0.14 & 89s \\
\hline
\multirow{1}{*}{QHMC-soft-ad} &  0.19 & 0.13  & 75s &{HMC-soft-ad} & 0.22 & 0.15 & 83s \\
\hline

\multirow{1}{*}{QHMC-var}  & 0.20 &  0.12 & 80s & {HMC-var} & 0.23 & 0.13 & 91s \\ \hline
\multirow{1}{*}{QHMC-soft-var}  &  0.21 & 0.13 & 71s & {HMC-soft-var} & 0.24 & 0.14 & 79s \\ \hline
\multirow{1}{*}{QHMC-both}  & \textbf{0.13} & 0.12 & 86s & {HMC-both}& 0.15 & 0.14 & 97s \\ \hline
\multirow{1}{*}{QHMC-soft-both}  & 0.14 & 0.13  & 74s & {HMC-soft-both} & 0.15 & 0.15 & 82s \\ \hline
\multirow{1}{*}{tnQHMC}  & 0.15 & 0.13 & 96s & {tnHMC} & 0.16 & 0.16 & 103s \\ \hline
\end{tabular}
\label{table: QHMCvsHMCsolute}
\end{center}
\end{table}

\subsubsection{Example 4: Heat Transfer in a Hallow Sphere}
\begin{figure}[ht]
\begin{center}
\includegraphics[width=9cm]{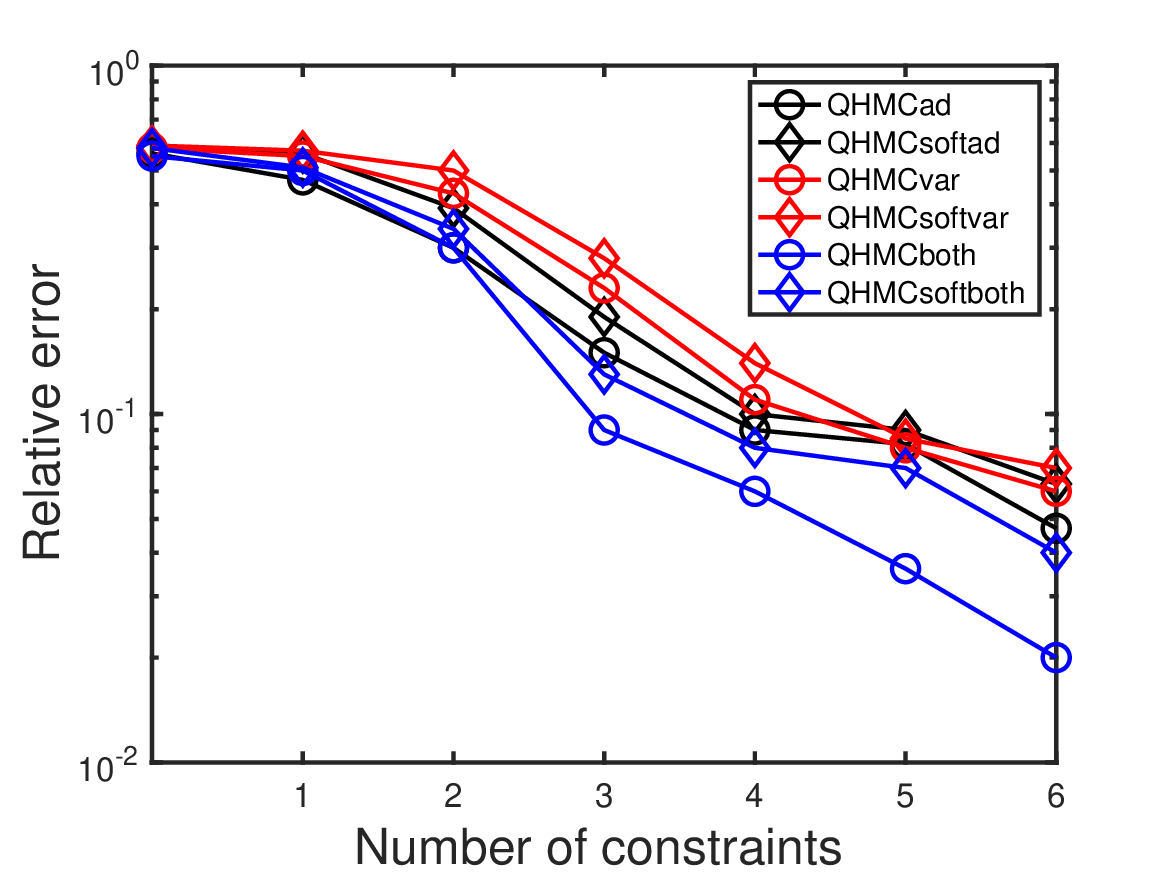}
\end{center}
\caption{The change in relative error while adding constraints, heat equation.}
\label{fig: numconstraints}
\end{figure}

\begin{figure}[htbp]
    \centering
    \begin{subfigure}[b]{0.49\textwidth}
        \includegraphics[width=\textwidth]{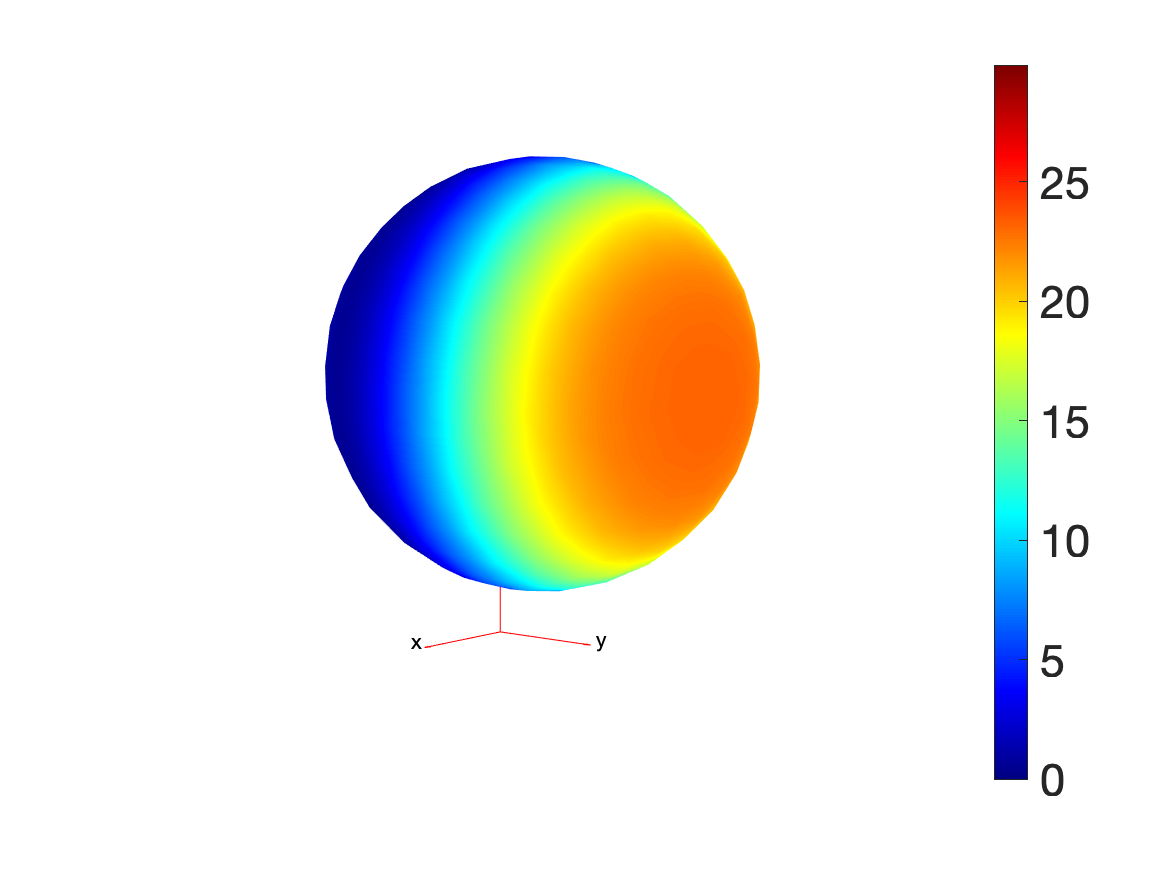}
        \caption{Heat equation data, ground truth $y(x)$.}
        \label{fig: ground3}
    \end{subfigure}
    \hfill
    \begin{subfigure}[b]{0.49\textwidth}
        \includegraphics[width=\textwidth]{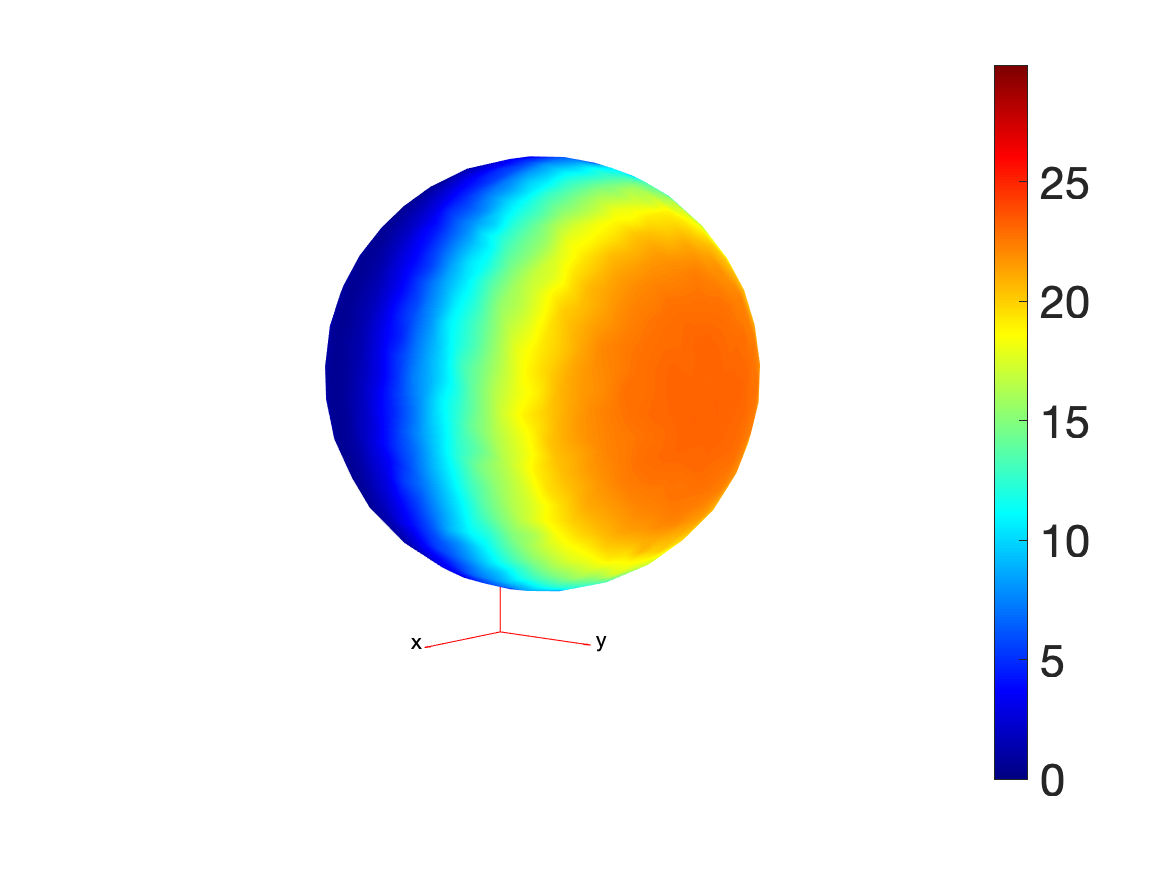}
        \caption{QHMCsoftboth prediction $y^{*}(x)$.}
        \label{fig: predd}
    \end{subfigure}
    \caption{Comparison of the ground truth and QHMCsoftboth result.}
    \label{fig: main2}
\end{figure}

In this 3-dimensional example, we consider a heat transfer problem in a hallow sphere. Let $B_{r}(0)$ represent a ball centered at $0$ with radius $r$. Defining the hallow sphere as $D = B_{4}(0) - B_{2}(0)$, the equations are given as~\citep{3Dyang2018}
\begin{align}
    \begin{cases}
    \frac{\partial u(\textbf{x}, t)}{\partial t} - \nabla \cdot (\kappa \nabla u(\textbf{x}, t)) = 0, & \textbf{x} \in D, \\
    \kappa \frac{\partial u(\textbf{x}, t)}{\partial \textbf{n}} = \theta^2(\pi - \theta)^2{\phi^2(\pi - \phi)^2}, & \text{if } \|\textbf{x}|_2 = 4 \text{ and } \phi \geq 0, \\
    \kappa \frac{\partial u(\textbf{x}, t)}{\partial \textbf{n}} = 0, & \text{if } \|\textbf{x}\|_2 = 4 \text{ and } \phi < 0, \\
u(\textbf{x}, t) = 0, & \text{if } \|\textbf{x}\|_2 = 2.
    \end{cases}
\end{align}
In this context, $\mathbf{n}$ denotes the normal vector pointing outward, while $\theta$ and $\phi$ represent the azimuthal and elevation angles, respectively, of points within the sphere. We determine the precise heat conductivity using $ \kappa = 1.0 + \exp(0.05u)$. The  quadratic elements with 12,854 degrees of freedom are employed, and we set $y(\mathbf{x}) = u(\mathbf{x}, 10)$ to solve the PDEs. Starting with 6 initial locations at 0 on the surface, 6 new constraint locations are introduced based on the active learning approach of the QHMC version. In Figure~\ref{fig: numconstraints}, we can observe the decrease in relative error while the constraints are added step by step. In addition, Figure~\ref{fig: main2} shows the ground truth and the GP result obtained by QHMCsoftboth algorithm, where we see that QHMCsoftboth $y^{*}(x)$ matches the reference model. Moreover, its posterior variance is small based on the results shown in Table~\ref{table: QHMCvsHMCheat}. The table provides the error, posterior variance and time performances of QHMC and HMC algorithms, and we can see the advantages of QHMC over HMC in all categories, even with the truncated Gaussian algorithm. Although all of the algorithms complete the GP regression in less than $1$ minute, comparing the truncated Gaussian method with QHMC-based algorithms, we observe $40-60\%$ time efficiency along with compatible accuracy of QHMC algorithms. In addition to the time and accuracy performances, it is shown that the posterior variance values are smallest with QHMCvar and QHMCboth approaches, followed by tnQHMC and QHMCad approaches. Using HMC sampling in all methods generates larger posterior variances. 

\begin{table}
\begin{center}
\caption[Comparison of QHMC and HMC on heat transfer with nonnegativity.]{Comparison of QHMC and HMC on heat transfer with nonnegativity.}
 \begin{tabular}{ |l || c | c| c| l ||c | c | c| c| r }
\hline
\hline 
 Method &  Error &  Posterior Var & Time  & Method &  Error &  Posterior Var & Time  \\ \hline \hline
\multirow{1}{*}{QHMC-ad} & 0.04 & 0.04 & 34s & {HMC-ad}& 0.06 & 0.07 & 40s \\
\hline
\multirow{1}{*}{QHMC-soft-ad} & 0.05  & 0.04  & 30s &{HMC-soft-ad} & 0.07 & 0.07 & 32s \\
\hline

\multirow{1}{*}{QHMC-var}  &  0.05 & 0.02  & 30s & {HMC-var} & 0.09 & 0.05 & 27s \\ \hline
\multirow{1}{*}{QHMC-soft-var}  & 0.06 & 0.03 & 26s & {HMC-soft-var} & 0.10 & 0.05 & 29s \\ \hline
\multirow{1}{*}{\textbf{QHMC-both}}  & \textbf{0.02} & 0.03 & 32s & {HMC-both}& 0.04 & 0.05 & 37s \\ \hline
\multirow{1}{*}{QHMC-soft-both}  & 0.03 & 0.03 & 27s & {HMC-soft-both} & 0.05 & 0.06 & 35s \\ \hline
\multirow{1}{*}{tnQHMC}  & 0.04 & 0.05 & 51s & {tnHMC} & 0.06 & 0.07 & 56s \\ \hline
\end{tabular}
\label{table: QHMCvsHMCheat}
\end{center}
\end{table}

\begin{figure}[htbp]
    \centering
    \begin{subfigure}[b]{0.49\textwidth}
        \includegraphics[width=\textwidth]{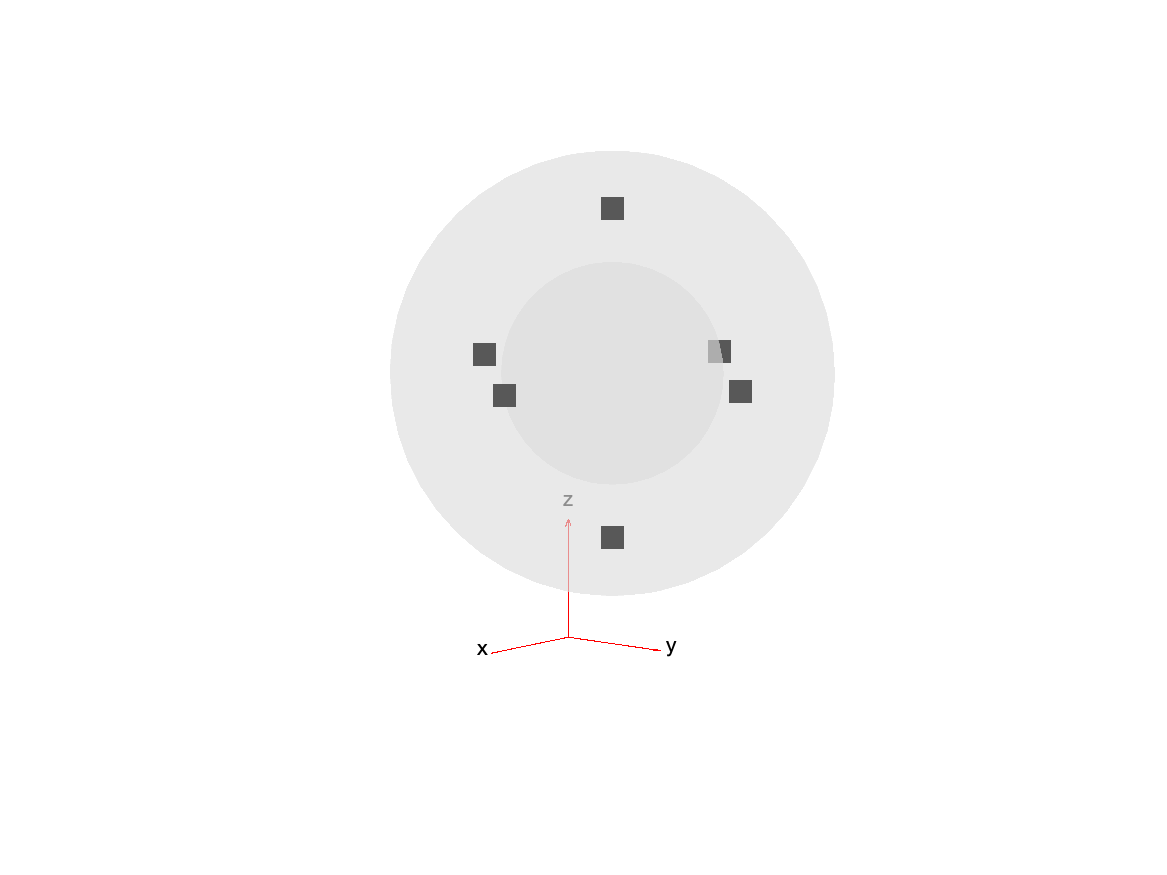}
        \caption{Initial locations}
        \label{fig: initialloc}
    \end{subfigure}
    \hfill
    \begin{subfigure}[b]{0.49\textwidth}
        \includegraphics[width=\textwidth]{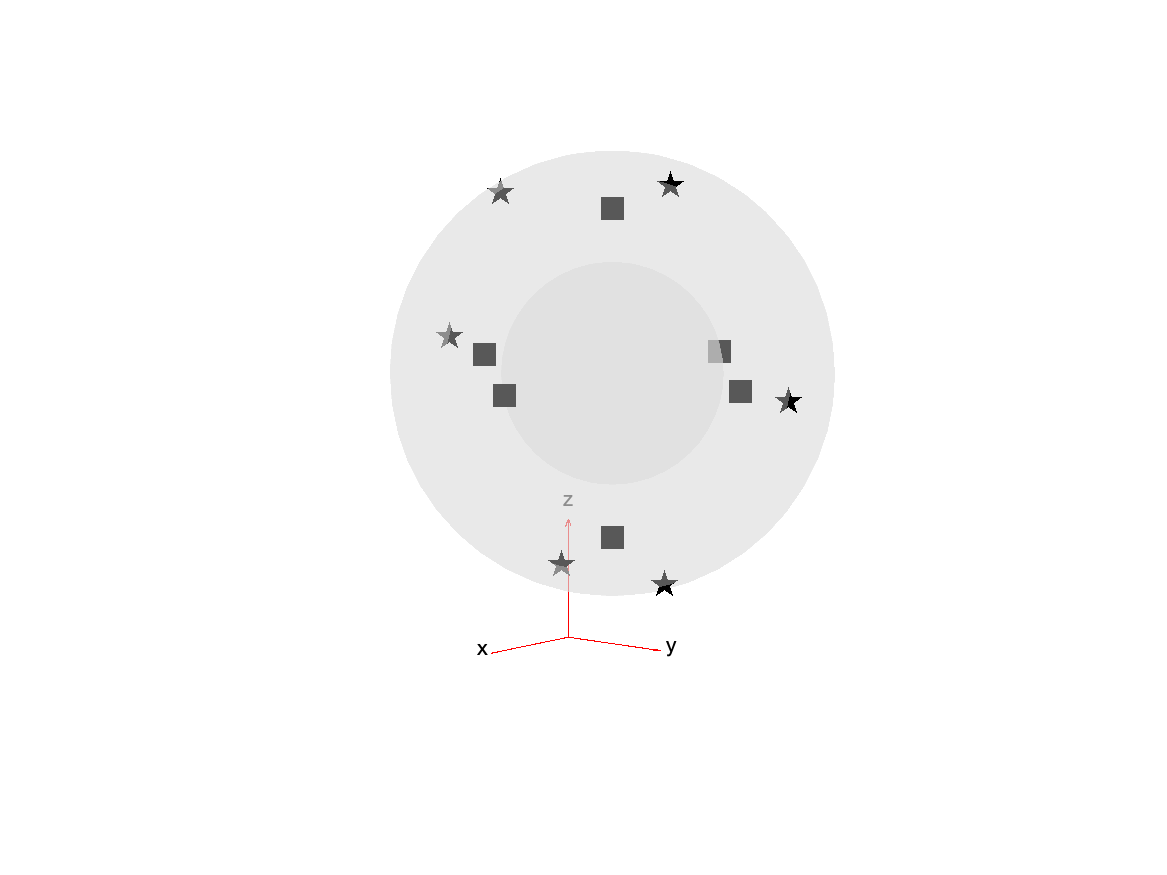}
        \caption{Constraint locations added by QHMC.}
        \label{fig: constraintloc}
    \end{subfigure}
    \caption{Initial locations and adaptively added constraint locations.}
    \label{fig: heatSteps}
\end{figure}
\subsection{Monotonicity Constraints}
In this section, we provide two numerical examples to investigate the effectiveness of our algorithms on monotonicity constraints. We enforce the monotonicity constraints in the direction of active variables. Similar to the comparisons in previous section, we illustrate the advantages of QHMC over HMC, and then compare the performance of QHMC algorithms with additive GP approach introduced in~\citep{lopezmonotonicity} with respect to the same criteria as in the previous section.
\subsubsection{Example 1}
We consider the following 5D function with monotonicity constraints~\citep{lopezmonotonicity}:
\begin{align*}
    f(x) = \arctan{(5x_{1})}+ \arctan{(2x_{2})} + x_{3} + 2x_{4}^{2} + \frac{2}{1+ \exp{-10(x_{5}-\frac{1}{2})}}.
\end{align*}

\begin{figure}[ht]
\begin{center}
\includegraphics[width=15cm]{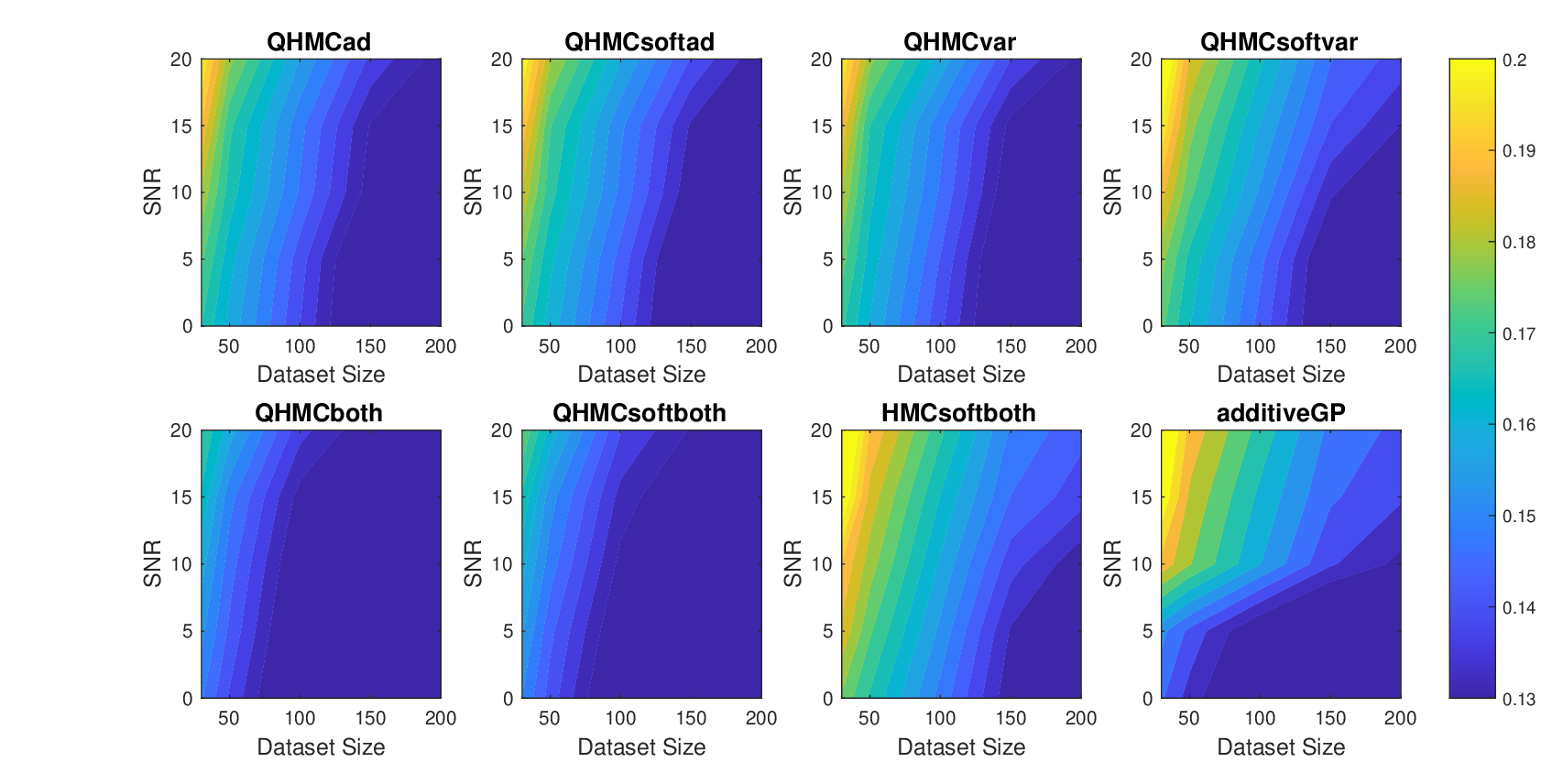}
\end{center}
\caption{Relative error of the algorithms with different data sizes and signal to noise ratios~(SNR) for Example 1~(5D), monotonicity.}
\label{fig: error5dMonotonicity}
\end{figure}

\begin{figure}[ht]
\begin{center}
\includegraphics[width=15cm]{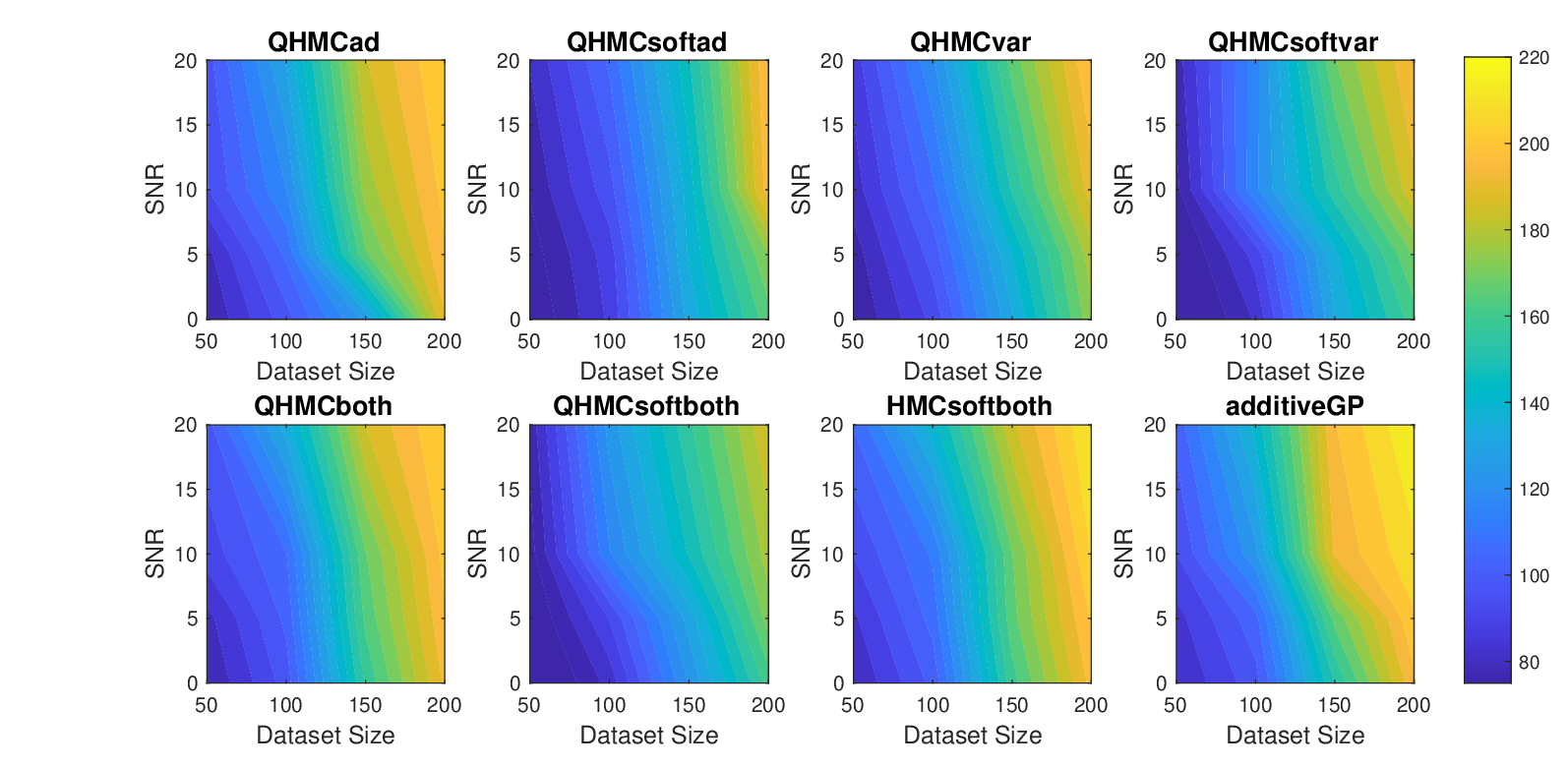}
\end{center}
\caption{Time comparison of the algorithms with different data sizes and signal to noise ratios~(SNR) for Example 1~(5D), monotonicity.}
\label{fig: time5dMonotonicity}
\end{figure}

Table~\ref{table: QHMCvsHMC5Dmonotonicity} shows the performances of HMC and QHMC algorithms, where we observe that QHMC achieves higher accuracy with lower variance in a shorter amount of time. The comparison proves that each version of QHMC is more efficient than HMC In addition, Figure~\ref{fig: error5dMonotonicity} shows the relative error values of QHMC and additive GP algorithms with respect to the change in SNR and dataset size. Based on the results, it is clear that QHMCboth and QHMCsoftboth provide the most accurate results under every different condition, while the difference is more remarkable for the cases in which noise is higher. Although QHMCboth and QHMCsoftboth provides the most accurate results, other QHMC versions also generate more accurate results then additive GP method. Moreover, Figure~\ref{fig: time5dMonotonicity} shows that the soft-constrained QHMC approaches are faster than the hard-constrained QHMC, while hard-constrained QHMC versions are still faster than additive GP algorithm. 

\begin{table}
\begin{center}
\caption[Comparison of QHMC and HMC on 5D, monotonicity.]{Comparison of QHMC and HMC on 5D, monotonicity.}
 \begin{tabular}{ |l || c | c| c| l ||c | c | c| c| r }
\hline
\hline 
 Method &  Error &  Posterior Var & Time & Method &  Error &  Posterior Var & Time  \\ \hline \hline
\multirow{1}{*}{QHMC-ad} & 0.11 & 0.16 & 2m 23s & {HMC-ad}& 0.13 & 0.17 & 3m 14s \\
\hline
\multirow{1}{*}{QHMC-soft-ad} & 0.14  & 0.18  & 1m 57s&{HMC-soft-ad} & 0.17 & 0.20 & 2m 48s \\
\hline

\multirow{1}{*}{QHMC-var}  &  0.12 & 0.15  & 2m 13s & {HMC-var} & 0.15 & 0.17 & 2m 58s \\ \hline
\multirow{1}{*}{QHMC-soft-var}  & 0.15 & 0.17 & 1m 42s & {HMC-soft-var} & 0.18 & 0.19 & 2m 16s \\ \hline
\multirow{1}{*}{\textbf{QHMC-both}}  & \textbf{0.10} & 0.13 & 2m 25s  & {HMC-both}& 0.12 & 0.15 & 2m 58s \\ \hline
\multirow{1}{*}{\textbf{QHMC-soft-both}}  & 0.12 & 0.14 & \textbf{1m 55s} & {HMC-soft-both} & 0.14 & 0.15 & 2m 39s\\ \hline
\end{tabular}
\label{table: QHMCvsHMC5Dmonotonicity}
\end{center}
\end{table}

\begin{figure}[ht]
\begin{center}
\includegraphics[width=15cm]{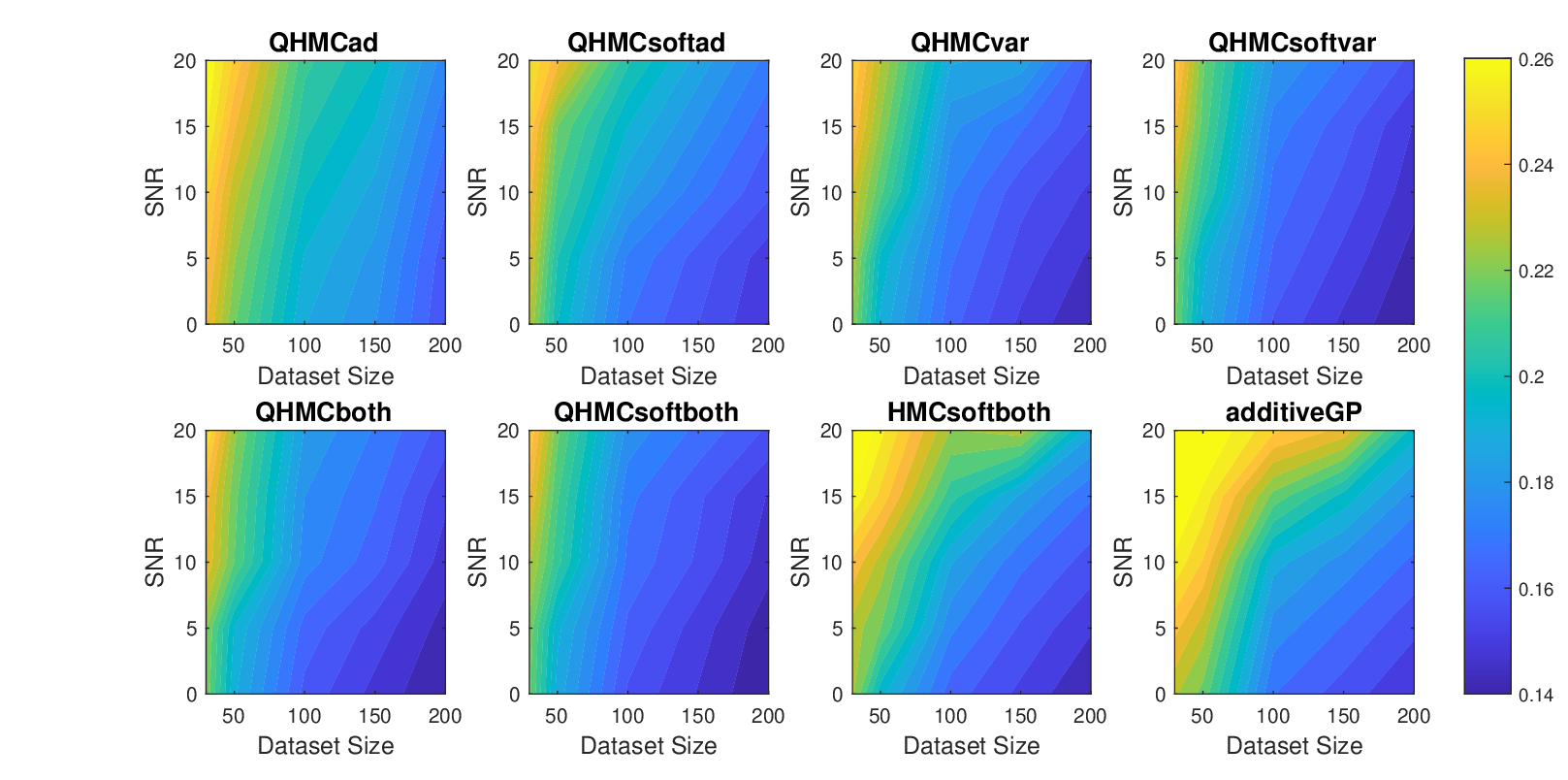}
\end{center}
\caption{Posterior variances of the algorithms with different data sizes and signal to noise ratios~(SNR) for Example 1~(5D), monotonicity.}
\label{fig: 5dMonotonicityVar}
\end{figure}
\subsubsection{Example 2}
\begin{figure}[ht]
\begin{center}
\includegraphics[width=15cm]{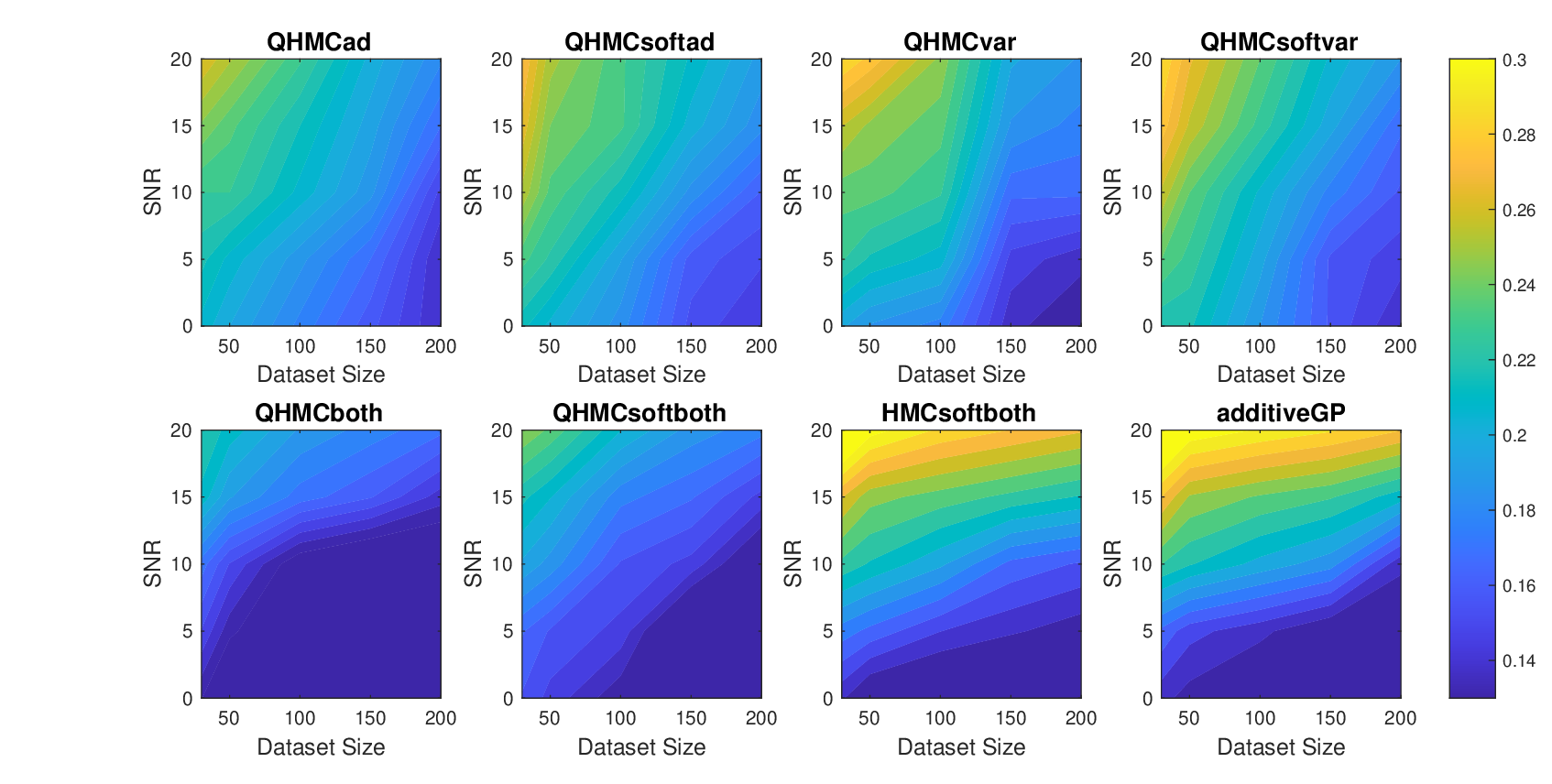}
\end{center}
\caption{Relative error of the algorithms with different data sizes and signal to noise ratios~(SNR) for Example 2~(20D), monotonicity.}
\label{fig: error20dMonotonicity}
\end{figure}

\begin{figure}[ht]
\begin{center}
\includegraphics[width=15cm]{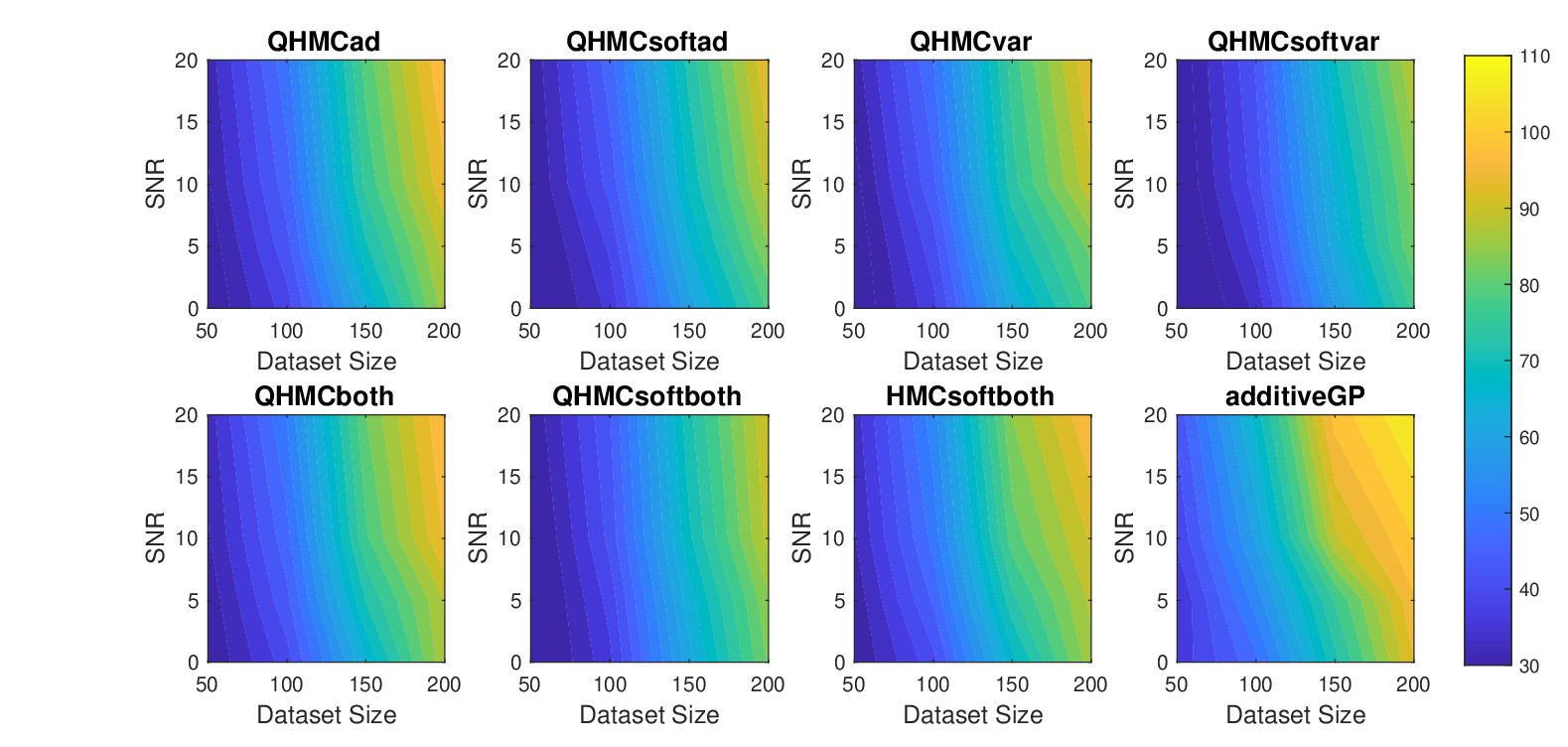}
\end{center}
\caption{Time comparison of the algorithms with different data sizes and signal to noise ratios~(SNR) for Example 2~(20D), monotonicity.}
\label{fig: time20dMonotonicity}
\end{figure}

\begin{figure}[ht]
\begin{center}
\includegraphics[width=15cm]{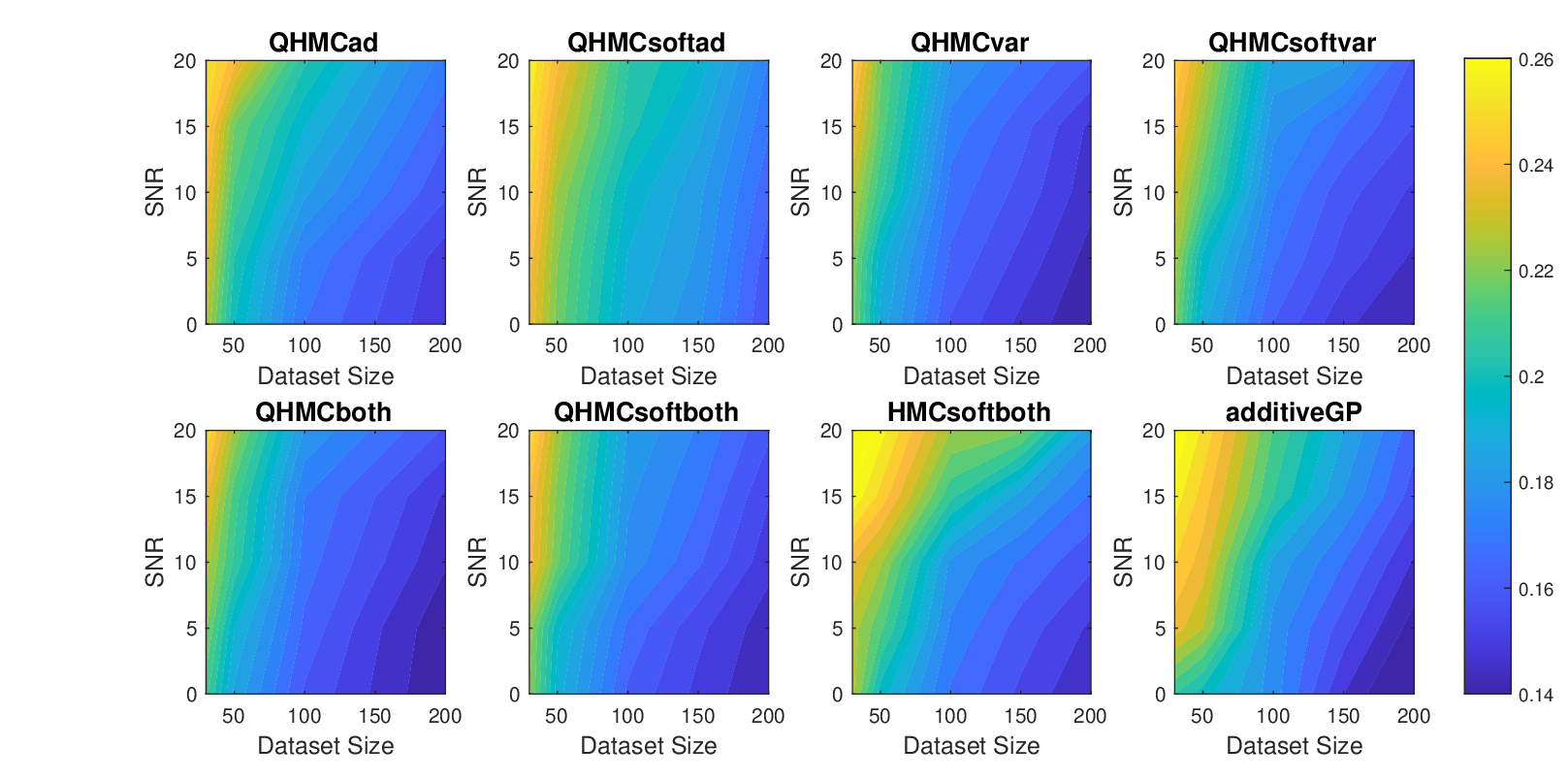}
\end{center}
\caption{Posterior variances of the algorithms with different data sizes and signal to noise ratios~(SNR) for Example 2~(20D), monotonicity.}
\label{fig: 20dMonotonicityVar}
\end{figure}

We consider the target function used in~\citep{lopezmonotonicity, lopezGPseq}
\begin{align*}
        f(x_{1},x_{2},...,x_{d}) = \sum_{i=1}^{d} 
        \arctan{5 \left [1 - \frac{i}{d+1} \right]x_{i}}, \quad \text{where} \quad d=2.
\end{align*}
In Table~\ref{table: QHMCvsHMC20Dmonotonicity}, we illustrate accuracy and time advantages of QHMC over HMC. For each version of QHMC and HMC, we see that using QHMC sampling in a specific version accelerates the process while increasing the accuracy. Overall comparison shows that among all versions with QHMC and HMC sampling, QHMCboth is the most accurate approach, while QHMCsoftboth is the fastest and ranked second in accuracy. In this set of experiments, we included the results of HMCsoftboth in the comparison of QHMC-based methods and additive GP. Figure~\ref{fig: error20dMonotonicity} and Figure~\ref{fig: time20dMonotonicity} show the relative error and time performances of QHMC-based algorithms, HMCsoftboth and additive GP algorithm, respectively. In this last example with the highest dimension, we observe the same phenomena as the previous results, in which soft-constrained versions are more efficient, while hard-constrained QHMC approaches are still faster than additive GP under various conditions such as high noise. Depending on Figure~\ref{fig: error20dMonotonicity} and Figure~\ref{fig: 20dMonotonicityVar}, we can state that QHMCboth can tolerate noise levels up to $10\%$ with the smallest error and posterior variance, and it can still provide good accuracy~(error is around $0.15$) even when the SNR is higher than $10\%$. It is also worth to mention that although the error values generated by HMCsoftboth and additiveGP are pretty close, HMCsoftboth performs faster than additiveGP, especially when the dataset is larger and noise level is higher.
\begin{table}
\begin{center}
\caption[Comparison of QHMC and HMC on 20D, monotonicity.]{Comparison of QHMC and HMC on 20D, monotonicity.}
 \begin{tabular}{ |l || c | c| c| l ||c | c | c| c| r }
\hline

\hline 
 Method &  Error &  Posterior Var & Time & Method &  Error &  Posterior Var & Time  \\ \hline \hline
\multirow{1}{*}{QHMC-ad} & 0.13 & 0.18 & 33m 1s & {HMC-ad} & 0.15 & 0.21 & 35m 38s \\
\hline
\multirow{1}{*}{QHMC-soft-ad} & 0.15  & 0.19  & 31m 21s & {HMC-soft-ad} & 0.18 & 0.22 & 33m 41s \\
\hline

\multirow{1}{*}{QHMC-var}  &  0.14 & 0.16  & 32m 53s & {HMC-var} & 0.17 & 0.17 & 34m 21s \\ \hline
\multirow{1}{*}{QHMC-soft-var}  & 0.16 & 0.17 & 29m 42s & {HMC-soft-var} & 0.19 & 0.18 & 31m 17s \\ \hline
\multirow{1}{*}{\textbf{QHMC-both}}  & \textbf{0.11} & 0.14 & 33m 45s & {HMC-both} & 0.14 & 0.16 & 36m 21s \\ \hline
\multirow{1}{*}{\textbf{QHMC-soft-both}}  & 0.12 & 0.15 & \textbf{29m 48s} & {HMC-soft-both}& 0.15 & 0.17 & 33m 11s \\ \hline
\end{tabular}
\label{table: QHMCvsHMC20Dmonotonicity}
\end{center}
\end{table}

\section{Conclusion and Discussion}
Leveraging the accuracy of QHMC training and the efficiency of probabilistic approach, we introduced a soft-constrained QHMC algorithm to enforce inequality and monotonicity constraints on the GP. The proposed algorithm reduces the difference between ground truth and the posterior mean in the resulting GP model, while increasing the efficiency by attaining the accurate results in a short amount of time. To further enhance the performance of the QHMC algorithms across various scenarios, we have implemented modified versions adopting adaptive learning. These versions provide flexibility in selecting the most suitable algorithm based on the specific priorities of a given problem.

We provided the convergence of QHMC by showing that its steady-state distribution approach the true posterior density, and theoretically justified that the probabilistic approach preserves convergence. Finally, we have implemented our methods to solve several types of optimization problems. In each experiment, we initially outlined the benefits of QHMC sampling in comparison to HMC sampling. These advantages remained consistent across all cases, resulting in approximately a $20\%$ time-saving and $15\%$ higher accuracy. 
Having demonstrated the advantages of QHMC sampling, we proceed to evaluate the performance of the algorithms across various scenarios. Our examples cover higher-dimensional problems featuring both inequality and monotonicity constraints. Furthermore, our evaluations include real-world applications where injecting physical properties is essential, particularly in cases involving inequality constraints. \\
In the context of inequality-constrained Gaussian processes (GPs), we explored 2-dimensional and 10-dimensional synthetic problems, along with two real applications involving 2-dimensional and 3-dimensional data. For synthetic examples, we observe the relative error, posterior variance and execution time of the algorithms while gradually increasing the noise level and dataset size. Overall, QHMC-based algorithms outperformed the truncated Gaussian methods. Although the truncated Gaussian methods provide high accuracy in the absence of noise and are compatible with QHMC approaches, their relative error and posterior variances increase as the noise appeared and increased. Moreover, the advantages of soft-constrained QHMC became more evident, particularly in higher-dimensional cases, when compared to truncated Gaussian and even hard-constrained QHMC. The time comparison of the algorithms underscores that the truncated Gaussian methods are significantly impacted by the curse of dimensionality and large datasets, exhibiting slower performance under these conditions. In real-world application scenarios featuring 2-dimensional and 3-dimensional data, the findings were consistent with those observed in the synthetic examples. Although the accuracy level may not reach the highest levels observed in the synthetic examples and 3-dimensional heat equation problem, the observed trend remains consistent. The lower accuracy observed in the latter problem can be attributed to the non-homogeneous structure of solute concentration. \\
In the case of monotonicity-constrained GP, we addressed 5-dimensional and 20-dimensional examples, utilizing the same configuration as employed for inequality-constrained GP. A comprehensive comparison was conducted between all versions of QHMC algorithms and the additive GP method. The results indicate that QHMC-based approaches hold a notable advantage, particularly in scenarios involving noise and large datasets. While additive GP proves to be a strong method suitable for high-dimensional cases, QHMC algorithms performed faster and yielded lower variances.\\
In conclusion, the work has demonstrated that soft-constrained QHMC is a robust, efficient and flexible method that can be applicable to higher dimensional cases and large datasets. Numerical results have shown that soft-constrained QHMC is promising to be generalized to various applications with different physical properties.
\bibliographystyle{plain}
\bibliography{test}  
\end{document}